# 4CNet: A Diffusion Approach to Map Prediction for Decentralized Multi-Robot Exploration

Aaron Hao Tan, *IEEE Student Member*, Siddarth Narasimhan, *IEEE Student Member,* Goldie Nejat, *IEEE Member*

***Abstract***— **Mobile robots in unknown cluttered environments with irregularly shaped obstacles often face energy and communication challenges which directly affect their ability to explore these environments. In this paper, we introduce a novel deep learning architecture, Confidence-Aware Contrastive Conditional Consistency Model (*4CNet*), for robot map prediction during decentralized, resource-limited multi-robot exploration. *4CNet* uniquely incorporates: 1) a conditional consistency model for map prediction in unstructured unknown regions, 2) a contrastive map-trajectory pretraining framework for a trajectory encoder that extracts spatial information from the trajectories of nearby robots during map prediction, and 3) a confidence network to measure the uncertainty of map prediction for effective exploration under resource constraints. We incorporate *4CNet* within our proposed robot exploration with map prediction architecture, *4CNet-E.* We then conduct extensive comparison studies with *4CNet-E* and state-of-the-art heuristic and learning methods to investigate both map prediction and exploration performance in environments consisting of irregularly shaped obstacles and uneven terrain. Results showed that *4CNet-E* obtained statistically significant higher prediction accuracy and area coverage with varying environment sizes, number of robots, energy budgets, and communication limitations when compared to database and learning-based methods. Hardware experiments were performed and validated the applicability and generalizability of *4CNet-E* in both unstructured indoor and real natural outdoor environments.**

## I. INTRODUCTION

MOBILE robots can be deployed in unknown and resource-limited environments to complete a variety of tasks, including searching for victims in disaster scenes [1]–[3], forest coverage [4], and planetary exploration [5], [6]. These environments feature irregularly shaped obstacles (i.e., flower beds, tree clusters, bushes) and uneven terrain (i.e., hills, rough ground surfaces), and can be

This work was supported in part by the Natural Sciences and Engineering Research Council of Canada (NSERC), and in part by the Canada Research Chairs program (CRC). *(Corresponding author: Aaron Hao Tan).*
The authors are with the Autonomous Systems and Biomechatronics Laboratory (ASBLab), Department of Mechanical and Industrial Engineering, University of Toronto, Toronto, ON M5S 3G8, Canada (e-mail: aaronhao.tan@utoronto.ca; s.narasimhan@mail.utoronto.ca; nejat@mie.utoronto.ca).

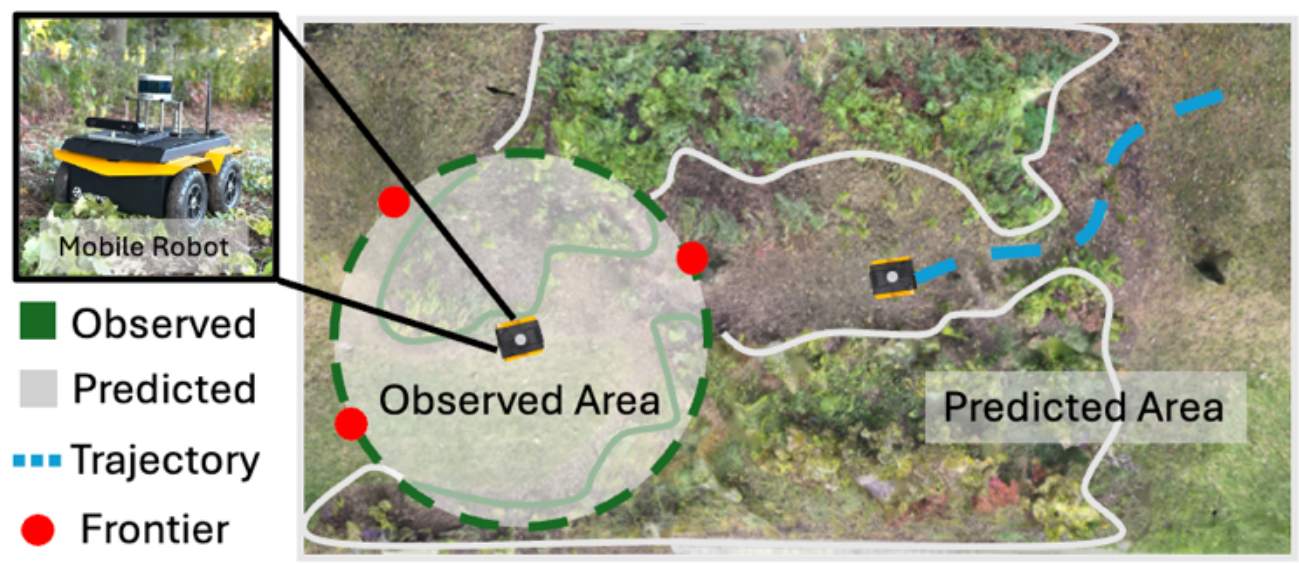

**Fig. 1.** A mobile robot equipped with *4CNet* predicts the unobserved map based on the observed area, and the trajectory of a nearby robot. The predicted map is used during frontier selection in an outdoor environment with irregularly shaped obstacles (bushes) and uneven terrain (small hills).

dynamic due to other robots pursuing their respective goals. Furthermore, mobile robots have limited onboard energy storage and communication capabilities [7]. For example, in disaster scenes, rubble can obstruct communication between robots, preventing the exchange of map information as this requires high-bandwidth connectivity. Similarly, in remote environments, the absence of charging stations and extreme conditions (i.e., harsh weather) can limit both energy usage and communication capabilities [8]–[10].

To explore an unknown environment, existing mobile robots mainly use frontier-based methods that directly estimate information gain from observed data based on distance and visibility to the frontiers [11], [12]. However, these methods naively label unobserved areas as either free-space or occupied [13], which can lead to suboptimal exploration and redundant efforts due to limited spatial awareness for the unobserved areas [14]. Robot map prediction addresses this limitation by estimating the spatial layout of unobserved regions, improving information gain estimation for frontier selection [15], Fig. 1., and enabling computation of additional navigation metrics such as traversability and energy expenditure. More importantly, in scenarios where mobile robots must navigate in uneven terrain while operating with limited energy budgets and communication capabilities, map prediction: 1) enables the anticipation of steep inclines, preventing excessive energy consumption [15], 2) results in reduced travel during exploration, promoting energy conservation [16], and 3) minimizes the need for high-bandwidth communication of robot maps with other robots, thus alleviating the communication burden [17].

Currently, map prediction methods are limited to structured, indoor environments with repetitive rectilinear features [18]–[23]. However, robot map prediction in unstructured, natural outdoor environments with irregularly shaped obstacles and uneven terrain is more complex due to non-repetitive environmental geometries and noisy sensory information. Additionally, existing map prediction methods have only been used in static single robot environments and have not been extended to consider dynamic multi-robot scenarios in which other robots are present and completing their own goals. However, the trajectories of these other robots can provide implicit information about the spatial layout of unobserved regions [24]. Specifically, a robot's deviation from a linear trajectory (indicating no direct route) characteristically implies the presence of an obstacle [25]. Therefore, a robot's trajectory within an unknown environment can reveal obstacle contours that have not been directly observed. Hence, robot trajectories offer a low-bandwidth approximation of robot maps in terms of providing spatial information while minimizing communication demand.

In this paper, we present a novel decentralized multi-robot exploration method called Confidence-Aware Contrastive Conditional Consistency Model (*4CNet*), to predict (*foresee*) unknown spatial configurations in unstructured multi-robot environments with irregularly shaped obstacles and uneven terrain. *4CNet* is incorporated within our proposed decentralized multi-robot exploration with map prediction architecture, *4CNet-E*, to efficiently explore unknown environments under limited energy budgets and communication constraints. Specifically, our contributions include:

1) The development of a novel map prediction network that is the first to utilize a consistency model to reconstruct unobserved spatial configurations in dynamic, unstructured, multi-robot environments.
2) A novel contrastive trajectory encoding framework that uniquely extracts spatial features from multi-robot trajectories to incorporate static environment features and dynamic robot trajectory features in communication-limited environments.
3) The first implementation of a confidence network for map prediction that quantifies prediction uncertainty, to guide energy-constrained robots towards high-uncertainty regions and maximize prediction accuracy.

## II. Related Works

The review presented include: 1) map prediction methods [14], [16], [18]–[23], [26]–[31], 2) energy-aware exploration methods [11], [32]–[42], and 3) diffusion methods for navigation [43]–[52].

### *A. Map Prediction Methods*

Map prediction methods can be categorized into heuristic-based [18]–[22] or deep learning (DL)-based [14], [16], [26]–[31] methods. Heuristic-based methods utilize predefined rules to interpret spatial layouts by exploiting structural patterns in the robot's environment. In contrast, DL methods utilize deep neural networks to extract spatial features within a high-dimensional space in order to make map predictions.

#### *1) Heuristic-based Methods*

Heuristic map prediction methods have used either map databases (DB) [18]–[20], representative lines [21], or low-rank matrix completion (LRMC) [22] to define a set of predefined rules that have interpreted spatial layouts for the exploitation of structural patterns in the robot's environment.

In DB methods, individual robots have access to a database of 2D robot maps prior to deployment. These maps are typically of repetitive structured environments such as rooms and corridors [19]. In general, DB methods have two phases [18]–[20]. Firstly, they identify a reference map from the database based on structural similarities to the unknown region using metrics such as feature vectors from Fast Appearance Based Mapping (FabMAP2) [19], counts of overlapping occupied cells [18], or highest likelihood of feature resemblance [20]. Secondly, they merge a reference map with the unknown region using techniques such as RANSAC-based Voronoi graph alignment [19], spatial alignment using homogenous transform matrices [18], or Gaussian filtering for spatial coherence [20].

Representative lines methods have been used in rectilinear environments, where 2D lines represent straight walls and corners [21]. These methods extrapolate lines from observed areas to unobserved regions using an objective function to maximize: 1) wall count uniformity by balancing distribution of walls in the predicted room layout, and 2) a simplicity index to minimize the shape complexity of predicted room layouts.

LRMC methods have been used to reconstruct missing cell information in a robot 2D map matrix by exploiting the low rank and incoherence characteristics of an environment [22]. These methods utilize an iterative Singular Value Decomposition solver to minimize the norm of the matrix by aggregating the singular values [53]. Missing cells in the map matrix are predicted by exploiting the linear relationships between the columns and rows of the map matrix.

The aforementioned map prediction methods have been applied to robot exploration [18]–[20], and path planning for coverage [21], [22] problems. Namely, for exploration, the predicted robot maps have been used to complete the exploration of unobserved areas [18] and improve frontier selection by providing expected information gain from unobserved parts of the environment [19], [20]. Experiments in structured indoor [18], [20] and repetitive underground tunnel environments [19] showed that exploration with map prediction provides more accurate maps and reduced travel distances when compared with non-map prediction exploration methods [54]. In coverage path planning, predicted maps inform offline methods such as Christofides [55] and tabu search [56]. Simulations in structured indoor [21] and grid world environments [22] have showed that path planning with map prediction has improved coverage ratio when compared to planning strategies without map prediction (i.e., lawnmower planning [57] and adaptive k-swap heuristic [58]).

#### *2) Deep Learning-based Methods*

DL based methods used for map prediction consist of autoencoder models [14], [16], [23], [28]–[31], or GANs [26], [27]. In an autoencoder model for robot map prediction, the encoder network uses down sampling layers to capture the spatial context from pixel-level features from either partially observed 2D maps [14], [23], [29]–[31] or RGB-D images from robot-centric viewpoints [16], [28]. The decoder network then reconstructs the spatial embeddings extracted by the encoder network into map predictions through up sampling layers. Existing autoencoder models include network components such as either: 1) convolutional layers [31], or fully connected layers with skip connections [16], [23], [28]–[30], in the encoder and decoder networks, or 2) a probabilistic latent space bottleneck [14] between the encoder and decoder networks.

In contrast, robot map prediction methods using GANs generate maps with a game-theoretic approach involving a generator and a discriminator convolutional network [26], [27]. The generator network learns to generate complete robot maps by replicating the distribution of robot maps that are present in a dataset. The discriminator network then evaluates the generated map predictions with maps in the dataset to provide an evaluation signal for the loss function of the generator during training. This adversarial process progressively refines the ability to produce robot maps [59].

The above DL based map prediction methods have been used for robot exploration [14], [16], [26], [27], [29]–[31], and semantic mapping [28] in simulated and real-world structured environments. For exploration, the predicted map from autoencoder models and GANs were used with information-theoretic frontier selection methods to improve exploration efficiency over exploration methods without map prediction in terms of travel distance [16], [29], [14], [27], coverage [14], [30], exploration time [31], and map accuracy [26]. For semantic mapping, the predicted map was used to segment and label different objects within the environment, in order to fill in missing geometric data within the map of a robot [28].

### B. *Energy-Aware Exploration Methods*

Existing robot energy-aware exploration methods have focused on traversing unknown regions in an environment to generate maps, while minimizing travel distances and operation times [11], [60]. Maps or intended exploration goals were communicated between nearby robots in a team in a decentralized manner to select complementary goals in order to reduce redundant coverage [36]. These approaches can be categorized into three categories: 1) utility-based [11], [39]–[42], 2) market-based [32]–[35], and 3) deep reinforcement learning (DRL) methods [36]–[38].

Utility-based robot exploration methods have used a utility function to select frontiers that maximize expected information gain and minimize travel cost [11], [39]–[42]. The objective is to maximize coverage and minimize robot energy consumption. On the other hand, market-based exploration methods have used a robot bidding procedure for frontiers based on estimated travel cost (distance, time) for exploration [32]–[35]. Individual robots minimize their travel cost, while the team maximizes its coverage. DRL techniques such as Double Deep Q Recurrent Networks (DDRQN) [36], Multi-agent Proximal Policy Optimization (MAPPO) [37], and Multi-agent Deep Deterministic Policy Gradient (MADDPG) [38], have been used to train decentralized exploration policies that maximize the expected discounted reward over the horizon of an episode. These reward functions positively reward a robot for area coverage while negatively reward distance traveled and time elapsed. Robot experiments were conducted in both simulation and real-world structured environments, to highlight the difference in travel distance/time between exploration methods.

### C. *Diffusion Methods for Navigation*

Diffusion models are a class of generative models that generate predictions by iteratively denoising Gaussian noise through a multi-pass process [61], [62]. Traditionally, diffusion models have been primarily used for various generative modeling tasks including image generation [63], [64], image super-resolution [65], [66], image inpainting [67], [68], and image editing [69], [70]. These applications are focused on static image synthesis, which are distinct from the dynamic data generation required in robot navigation tasks. To the best of the authors' knowledge, Denoising Diffusion Probabilistic Models (DDPM) [71], and Diffusion Probabilistic Implicit Models (DDIM) [64], are the only diffusion models that has been applied to mobile robot navigation tasks to either predict robot velocities [44], [45], [48], [50], or waypoints [43], [46], [47], [49], [51], [52]. The DDPM architectures that have been used include U-Net [43]–[49], custom autoencoders [50], convolutional neural networks (CNNs) [51], [52]. The inputs to these models typically consist of sensory data such as RGB images [43], [44], [51], [52], LiDAR [48], [50], odometry [45], [49], and robot maps [46], [47], [51]. They are trained using navigation datasets consisting of ground truth robot trajectories, such as GNM [43], SACSoN [43], ETH pedestrian dataset [44], UCY pedestrian dataset [44], Maze2D [47], ImgEnv [48], and custom generated datasets in maze-like [50]–[52], or open space [45], [46], [49] environments. Evaluation of diffusion methods for navigation has been performed in both simulation [44]–[49], [51], [52], and real-world [43], [44], [46], [48], [50], [51], environments. In simulation, the proposed methods are tested in a synthetic maze simulator [47]–[49], [52], Trajdata simulator [44], OpenAI gym [45], or using offline datasets such as MRPB [51] and Gibson [46]. On the other hand, real-world evaluations include indoor [43], [46], [48], [50], [52], outdoor [43], [44], [46], and maze-like [51] environments, using robots such as LoCoBot [43], Agilex Hunter [50], Boston Dynamics Spot [51], or Unitree Go1 [51], [52]. Results from these experiments demonstrated that diffusion methods can successfully generate robot control commands (velocity, waypoints), outperforming existing navigation methods such as random walk, A* [72], and velocity-obstacle method [73] in terms of metrics such as success rate, number of collisions, navigation time, and success weighted by path length (SPL).

### D. *Summary of Limitations*

Existing heuristic and DL-based map prediction methods can suffer from lower prediction accuracy in terms of spatial structural features and pixel-level image feature textures [29]. Namely, heuristic methods assume similar static environments with rectilinear features, limiting their accuracy in unstructured settings with irregularly shaped obstacles and uneven terrain [18]–[22]. Meanwhile, DL models like autoencoders and GANs rely on single-pass through its network to generate predictions, which restricts their ability to capture complex spatial features, often resulting in blurry maps with inconsistent structural details [74]. GANs also face challenges with unstable training and mode collapse due to adversarial dynamics, which hinders their ability to represent diverse and complex environmental layouts [75], [76]. Existing map predictions methods also assume uniform uncertainty across all predicted pixels, which can lead to suboptimal decision-making during robot planning [77]. Furthermore, they also only focus on static single robot environments, and do not consider scenarios with multiple robots. In such scenarios, the trajectories of nearby robots can be uniquely leveraged to provide spatial context for unobserved regions [17].

In terms of existing energy-aware exploration methods, they do not address challenges where a robot: 1) do not have sufficient energy to explore an entire environment, and 2) cannot share map data due to communication limitations. Therefore, a robot exploration method is needed that can utilize map prediction to maximize coverage, while addressing limitations in onboard energy storage and communication capabilities.

Lastly, while diffusion models using the DDPM and DDIM variants have been applied to mobile robot navigation for generating low-resolution data, such as waypoints and velocities, they have not been used for high-resolution tasks like map prediction. This is due to the fact that map prediction requires generating 2D high-fidelity images of future environment states, which is difficult for DDPM and DDIMs as they incur significant computation cost and prediction time due to its training setup [64], [71]. As a result, this limitation motivates the use of faster diffusion model variants, such as the recently introduced consistency model [78], to generate detailed maps while reducing computational cost and prediction time during online robot exploration.

## III. The Robot Map Prediction Problem For Resource Limited Exploration

The robot map prediction problem for resource-limited exploration requires a mobile robot to explore an unknown, dynamic, and unstructured environment. The environment is dynamic, as it consists of other mobile robots achieving their own goals. Each robot explores the environment in a decentralized manner and operates under constrained communication capabilities and a limited energy budget. The goal is to maximize a robot's knowledge of the configuration of the environment using both the observed region of the environment, $M_i^{obs}$, and the predicted spatial configuration from the unobserved region of the environment, $M_i^{pred}$.

**Robots**: *N* number of non-holonomic mobile robots can exist in an environment, $R = \{r_1, r_2, \dots, r_n\}$. Each robot $r_i$ has an onboard sensor with a sensing range of $s$ for mapping its surroundings. The robots have their own individual exploration goals and only share their trajectory information when within $s$. Each $r_i$ also navigates at a fixed velocity, using a shortest path planner to generate feasibly trajectories between the robot's current position and the exploration goal. Energy consumption, $E_i$, for robot $r_i$ is modeled as a linearly decaying function with respect to the distance $\Delta d_i$ traversed by the robot, and the change in elevation $\Delta z_i$, at each timestep $t$, during robot navigation. Therefore, the energy consumption model at each time step $t$ is defined as:

$$E_i(\Delta d_{i,t}, \Delta z_{i,t}) = \left[ \left( w_i \cdot \Delta d_{i,t} \right) + \left( k_i \cdot \max(0, \Delta z_{i,t}) \right) \right], \tag{1}$$

where $w_i$ is the energy consumed per unit distance traveled, and $k_i$ is the energy cost for vertical motion along the *z*-axis. The function $\max(0, \Delta z_i)$ signifies that additional energy is consumed only when the robot is ascending ($\Delta z_i > 0$), while no additional energy is consumed for descending or traversing over flat terrain ($\Delta z_i \leq 0$). The trajectory of robot $r_i$ is denoted as $\tau_i$. This trajectory is a temporally ordered set of robot positions up to the current time step, $t$, and is represented as:

$$\tau_i = \{(x_{t_1}, y_{t_1}), (x_{t_2}, y_{t_2}), \dots, (x_t, y_t)\}. \tag{2}$$

**Environment**: The environment is represented by both traversable uneven terrain and non-traversable irregularly shaped obstacles. The environment is discretized into a heightmap $M_h$, where each cell in the map, $m_{(x,y)}$, is characterized by its $(x, y)$ coordinates and contains elevation information within a specified range $[l_{min}, l_{max}]$. Thus, the heightmap is defined as:

$$M_h = \{m_{(x,y)} \mid m_{(x,y)} \in [l_{min}, l_{max}], \forall (x, y)\}, \tag{3}$$

where each robot's position, $p_i = (x_i, y_i)$, represents a specific cell $m_{(x_i, y_i)}$ in $M_h$.

**Communication between Robots**: Robots can exchange their trajectories, $\tau_i$, when they are within sensing range, $s$, in a decentralized manner. The number of robots within $s$ is defined as $R_s; R_s \subseteq R$. To account for the stochastic nature of communication in real-world environments, a Communication Success Probability (CSP), $C \in [0,1]$, [36], is incorporated to represent the likelihood of successful trajectory transmissions between robots. To implement CSP, each robot position at time $t$ in the trajectory, $\tau_i^t$, is associated with a Bernoulli event represented by a random variable $B_t$. $B_t$ captures the success of communication: a value of 1 occurring with probability $q$, denotes a successful transmission of $\tau_i^t$; while a value of 0 with the complementary probability $1 - q$, denotes a transmission failure. The transmitted trajectory, $\tau_i'$, is a subset of the robot's traversed trajectory $\tau_i$. Specifically, $\tau_i'$ includes only those positions for which $B_t = 1$, and is expressed as:

$$\tau_i' = \{(x_t, y_t) \mid (x_t, y_t) \in \tau_i, B_t = 1\}. \tag{4}$$

Once $r_i$ receives trajectory information from all other robots within sensing range, $R_s$, the communicated trajectories are aggregated into a collective set $\delta_t = \{\tau_i'\}_{i \in R_s}$. We assume that each robot is able to obtain its own position with respect to a global frame acquired from an onboard global position system (GPS).

**Map Prediction Task:** Each robot predicts the unexplored environment configuration based on 1) its own observed portion of the environment during exploration, $M_i^{obs}$, and 2) the trajectory information of nearby robots, $\delta_t$. Thus, the predicted map $\widehat{M_\iota}$ of the entire environment is represented as:

$$\widehat{M_\iota} = f_\theta(\{M_i^{obs}, \delta_t\}). \tag{5}$$

$\widehat{M_\iota}$ is defined by $M_i^{obs} \cup M_i^{pred}$. The goal is to approximate the map prediction function $f_\theta$ such that $\widehat{M_\iota}$ can be used by a frontier-based exploration method, $\partial$, to account for both observed and predicted map information during exploration.

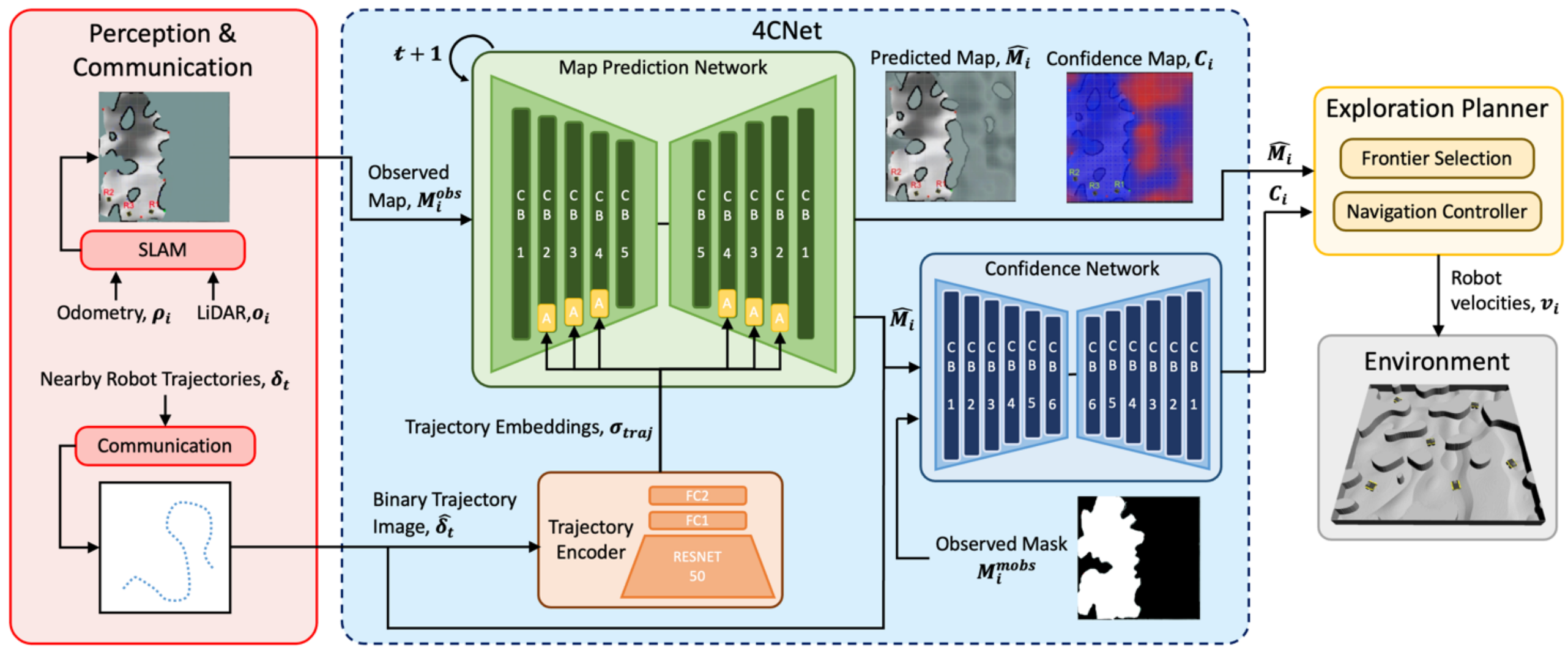

**Fig. 2.** The decentralized multi-robot exploration architecture, *4CNet-E,* for resource-limited exploration in unstructured environments with irregularly shaped obstacles and uneven terrain. **A** in the *Map Prediction Network* denotes the cross-attention mechanism for robot trajectory conditioning. **CB** represent Convolutional Blocks, and **FC** represent Fully Connected layers.

**Exploration Objective Function:** The exploration objective is to maximize robot spatial knowledge given a limited energy budget, $q_i$. Namely, the objective is to maximize the utility of a frontier, $U$, selected by $\partial$, over the time horizon, $h$, while adhering to the energy consumption model, $E_i$, of the robot $r_i$:

$$\text{maximize} \left[ \sum_{t=1}^{h} U\left(\partial\left(f_\theta\left(M_{i,t}^{obs}, \delta_t\right)\right)\right) \right], \qquad (6)$$
$$s.t. \sum_{t=1}^{h} E_i\left(\Delta d_{i,t}, \Delta z_{i,t}\right) \leq q_i.$$

## IV. The Robot Exploration Architecture with Map Prediction

The proposed DL robot exploration with map prediction architecture, *4CNet-E*, Fig. 2, has been developed to predict maps of partially explored environments consisting of irregular shaped obstacles and uneven terrain. The goal of this architecture is to use the predicted maps to guide robots towards unexplored regions with high expected information gain, while addressing fixed energy budgets and limited communication. It consists of three main subsystems: 1) *Perception and Communication,* 2) *4CNet*, and 3) *Exploration Planner. The Perception and Communication* subsystem generates a 2D partial map, $M_i^{obs}$, using robot LiDAR and odometry information via the *Simultaneous Mapping and Localization (SLAM)* module. It also communicates and receives trajectory information, $\delta_t$, from nearby robots using the *Communication* module. $M_i^{obs}$ and $\delta_t$ are used by the *Trajectory Encoder* module in *4CNet* to produce trajectory embeddings, $\sigma_{traj}$, that extract spatial features from $\delta_t$ to condition the *Map Prediction Network* (*MPN*). The *MPN* uses these embeddings to generate the predicted map $\widehat{M}_i$. $\widehat{M}_i$ is used by the *Confidence Network* (*CN*) to obtain a confidence map $C_i$ for each predicted pixel within $\widehat{M}_i$. The output of *4CNet* is both $\widehat{M}_i$ and $C_i$, which are used in the *Exploration Planner* subsystem to select a frontier goal location for the robot using a utility function $U$. The selected frontier goal is used by the *Navigation Controller* to generate a navigation trajectory, $\tau_{goal}$. The main subsystems of *4CNet-E* are discussed below in further details.

### *A. Perception and Communication Subsystem*

The SLAM module in the *Perception and Communication subsystem* utilizes robot odometry, $\rho_i$, and onboard 3D LiDAR observations, $o_i$, to localize the robot in the environment, $p_i$, and to generate a map of the observed regions, $M_i^{obs}$. Real-time Appearance Based Mapping (RTAB-Map) [79] is used to generate a graph-based map with nodes representing LiDAR scans. These nodes are connected by edges that represent spatial relationships between successive LiDAR scans. The *Communication* module is used to transmit the robot's own trajectory $\tau_i'$, and receive trajectory data, $\delta_t$, from other robots using coordinates acquired from the GPS. This is to enable robots to understand their relative positions within a global reference frame. The transmission of robot trajectories occurs when robots are within $s$. The transmitted $\delta_t$ are then transformed into the local reference frame of each robot's map, and represented as a 2D binary array, $\hat{\delta}_t$, that contains the composite trajectories of all nearby robots. Both $\hat{\delta}_t$, and $M_i^{obs}$ are provided to *4CNet* for map prediction.

### *B. 4CNet*

*4CNet* consists of three modules: 1) *Trajectory Encoder*, 2) *Map Prediction Network*, and 3) *Confidence Network.*

#### *1) Trajectory Encoder*

The objective of the *Trajectory Encoder* module is to encode the available trajectories from other robots, $\delta_t$, into a trajectory embedding vector, $\sigma_{traj}$. To train the *Trajectory Encoder*, we introduce a unique *Contrastive Map Trajectory Pretraining (CMTP)* framework to contrast robot trajectory and map features in order to capture spatial information (i.e., obstacle contours) implicitly present in robot trajectories. The *CMTP* framework utilizes two distinct encoders: a map encoder, $E_{map}$, for extracting spatial features from $M_h$ used only during pretraining, and a trajectory encoder, $E_{traj}$, for extracting robot coordinate features from $\widehat{\delta_t}$, used during both pretraining and inference. Both encoders use a ResNet50 backbone [80], and a projection head with two fully connected (FC) layers, Fig. 2. During contrastive learning [81], $E_{map}$ and $E_{traj}$ are pretrained simultaneously to align robot spatial and coordinate features from $M_h$ and $\widehat{\delta_t}$, respectively, into a common representation space, $\mathcal{R}$. Specifically, $E_{map}$ transforms the ground truth heightmap $M_h$ into a map embedding vector, $\sigma_{map}$. Concurrently, the trajectory image, $\widehat{\delta_t}$, is used by $E_{traj}$ to generate a robot trajectory embedding vector, $\sigma_{traj}$. The contrastive loss function, $\mathcal{L}_{CMPT}$, for *CMTP* is defined as:

$$\mathcal{L}_{CMTP} = -\mathbb{E}\left[\log \frac{\exp\left(\mathscr{k}\left(\sigma_{map}, \sigma_{traj}\right)/\mu\right)}{\sum_{\sigma'_{traj} \in \mathcal{D}_{MT}} \exp\left(\mathscr{k}\left(\sigma_{map}, \sigma'_{traj}\right)/\mu\right)}\right], \tag{7}$$

where $\mathcal{L}_{CMTP}$ computes the expected value of the negative log probability that a robot heightmap embedding, $\sigma_{map}$, has a higher similarity score with its corresponding trajectory embedding, $\sigma_{traj}$, compared to other trajectory embeddings, $\sigma'_{traj}$, within the dataset, $\mathcal{D}_{MT}$. Herein, $\mathcal{D}_{MT}$, is a dataset of robot heightmaps and trajectories. The function, $\mathscr{k}$, measures the cosine similarity between embedding pairs, outputting a similarity score. The exponential function, exp, is used to normalize similarity scores for effective gradient descent optimization during training. The temperature parameter, $\mu$, is used to control the sharpness of the distribution of similarity scores to directly influence the gradient magnitudes during backpropagation. Minimizing $\mathcal{L}_{CMTP}$ will maximize the similarity scores between map-trajectory embedding pairs. The *Trajectory Encoder* provides trajectory embeddings, $\sigma_{traj}$, to the *Map Prediction Network.*

*2) Map Prediction Network*

The *MPN* predicts the spatial configuration of an unobserved region within a partially explored environment using: 1) the robot observed map $M_i^{obs}$, and 2) the trajectory embeddings $\sigma_{traj}$. The *MPN* is designed as a 2D U-Net architecture, where the encoder consists of five convolutional blocks (CBs) with output channels of [128, 128, 256, 256, 512], Fig. 2. The decoder mirrors the encoder by having an equal number of convolutional blocks with the reverse output channels [512, 256, 256, 128, 128]. Each CB consists of a convolutional layer, batch normalization and ReLU activation. Cross-attention is integrated into the middle three blocks of both the encoder and decoder in the *MPN* to only integrate relevant trajectory features during the map prediction process. The $M_i^{obs}$ is represented as 224 × 224 pixel image. Environments of varying sizes are down sampled or padded accordingly to fit this resolution, to ensure that all input map is standardized while preserving the spatial characteristics of the original map.

The *MPN*'s prediction process follows the consistency model set up, which describe a variant of diffusion models designed for faster inference [78]. Specifically, consistency models require a two-stage approach for map prediction: a noising stage, and a denoising stage. In the **noising stage**, a Probability Flow Ordinary Differential Equation (PF ODE) is used to perturb the initial noiseless map state, $\mathfrak{m}_0$, to a terminal map state representing Gaussian noise, $\mathfrak{m}_T$, across time steps $t$. Each time step applies Gaussian noise according to a predefined schedule, generating a noisy map sequence $\{\mathfrak{m}_t\}_{t\in[\epsilon,T]}$, Fig. 3, where each step represents a map with gradual increase in noise. In the **denoising stage**, the *MPN*, $f_\theta$, reconstructs the original map $\mathfrak{m}_0$ from any noisy map state $\mathfrak{m}_t$ in the PF ODE sequence $\{\mathfrak{m}_t\}_{t\in[\epsilon,T]}$. Namely, the $f_\theta$ considers the time step of the sequence $t$, the partially known map $M_i^{obs}$, and the given robot trajectories $\delta_t$:

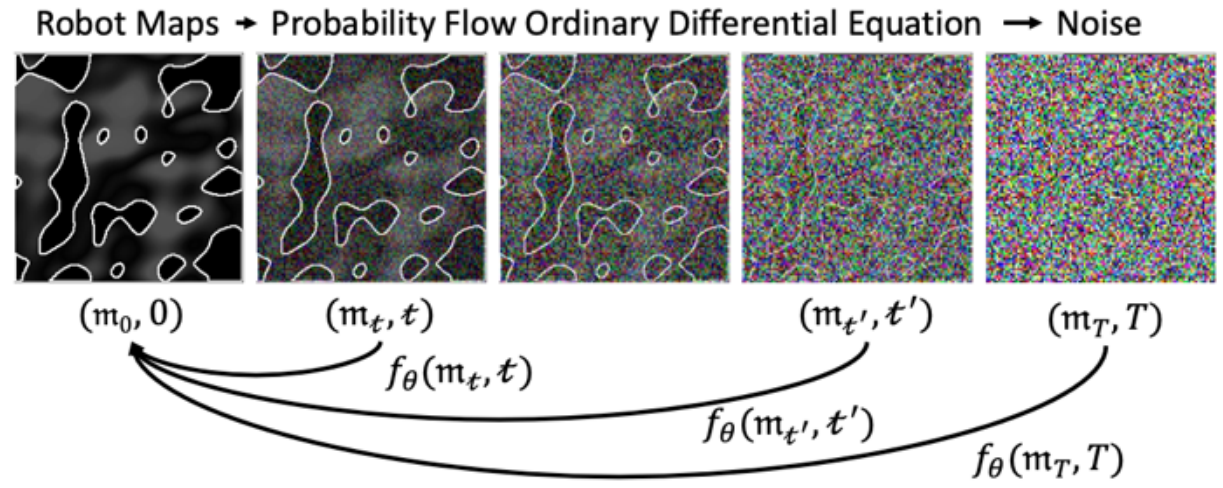

**Fig. 3.** Map prediction function, $f_\theta$, described as a consistency model that learns to transform any robot heightmap state, $\mathfrak{m}_t$, along the Probability Flow Ordinary Differential Equation sequence to the original map state, $\mathfrak{m}_0$. In the original map state at $\mathfrak{m}_0$, white lines represent obstacle contours, and the gray scale gradient represents traversable uneven terrain. $\mathfrak{m}_T$ represents the terminal map state consisting of Gaussian noise.

$$f_\theta\left(M_i^{obs}, t, \delta_t\right) = p_{\text{skip}}(t) M_h + p_{\text{out}}(t) F_\theta\left(M_i^{obs}, t, \delta_t\right), \tag{8}$$

where $p_{\text{skip}}$ and $p_{\text{out}}$ are weighting coefficients that serve two purposes. First, given a time step near the origin of the PF ODE, denoted by $t < \epsilon$, $f_\theta$ provides the ground truth heightmap, $M_h$, as the output. Second, as $t$ increases, the output of $f_\theta$ increasingly transitions to the output of the predictive model, $F_\theta$. The change in output between $M_h$ and $F_\theta$ is achieved by enabling $p_{\text{skip}}(t)$ to decrease and $p_{\text{out}}(t)$ to increase as $t$ increases. The initial conditions are $p_{\text{skip}}(\epsilon) = 1$ and $p_{\text{out}}(\epsilon) = 0$. Thus, $p_{\text{skip}}$ and $p_{\text{out}}$ ensure that $f_\theta$ is differentiable for model training through backpropagation. During inference, $f_\theta$ can be sampled with multiple passes through the map prediction network to refine the prediction quality iteratively.

For training of the *MPN*, an online network, $\boldsymbol{f_\theta}$, and a target network, $\boldsymbol{f_{\theta^-}}$, are used, where $\boldsymbol{f_\theta}$ is used to predict the robot heightmap state at $t + 1$, and $\boldsymbol{f_{\theta^-}}$ is used to predict the robot heightmap state at $t$, along the PF ODE sequence [82]. Note, the online network describes a network that is actively being updated to predict the robot heightmap state, whereas the target network is a non-updating network used to provide a stabilized reference for computing the *MPN* loss function, $\mathcal{L}_{MPN}$. To ensure stable training, the weights of $\boldsymbol{f_{\theta^-}}$ are set as an exponential moving average (EMA) of the weights of $\boldsymbol{f_\theta}$, described by:

$$\theta^- = \mathcal{H}\theta^- + (1 - \mathcal{H})\theta, \tag{9}$$

where $\mathcal{H}$ is the smoothing factor. EMA is used for $\theta^-$ to prevent drastic shifts in the $\boldsymbol{f_{\theta^-}}$ behavior due to large updates to the network parameters [82]. The $\mathcal{L}_{MPN}$ uses stochastic gradient descent to minimize the prediction differences between the two networks:

$$\mathcal{L}_{MPN} = \mathbb{E}\left[\lambda(t_n)\mathcal{L}_m\begin{pmatrix} f_\theta\left(\mathfrak{m}_{t_{n+1}} + t_{n+1}\xi, t_{n+1}, \tau_i\right), \\ f_{\theta^-}\left(\mathfrak{m}_{t_n} + t_n\xi, t_n, \tau_i\right) \end{pmatrix}\right]. \tag{10}$$

Herein, $\mathcal{L}_m$ computes the difference between the predicted maps from $f_\theta$ and $f_{\theta^-}$, while adhering to the consistency property in Eq. (8). The weighting function $\lambda(t_n)$ adjusts the significance of each term in the loss. $\xi$ is the Gaussian noise vector and represents the normally distributed noise sampled from $N(0, \mathrm{I})$ [78]. We designed $\mathcal{L}_m$ as a compound loss function that uses Perceptual Image Patch Similarity (LPIPS) [83] and the Edge Loss function, $\mathcal{L}_e$. In particular, LPIPS is used to evaluate the perceptual similarity score between the terrain features in the predicted maps, $(\mathfrak{m}_\theta, \mathfrak{m}_{\theta^-})$, from $f_\theta$ and $f_{\theta^-}$, respectively. The goal is to ensure that both predicted maps have high similarity in terms of visual and structural characteristics. The LPIPS function, $\mathrm{LPIPS}(\mathfrak{m}_\theta, \mathfrak{m}_{\theta^-})$, is expressed as [83]:

$$\mathrm{LPIPS}(\mathfrak{m}_\theta, \mathfrak{m}_{\theta^-}) = \phi\left(\left\|V\left(\mathfrak{m}_{\theta_i}\right) - V\left(\mathfrak{m}_{\theta^-_i}\right)\right\|_2^2\right), \tag{11}$$

where $V$ represents a function that extracts pixel features from the predicted maps, $\mathfrak{m}_\theta, \mathfrak{m}_{\theta^-}$. The function, $\phi$, applies a nonlinear transformation to the squared Euclidean distance between these extracted pixel features. This transformation converts the differences in the pixel feature space into the perceptual similarity score [83].

The Edge Loss function, $\mathcal{L}_e$, is used to compare boundary contours of irregularly shaped obstacles within the predicted robot heightmaps $(\mathfrak{m}_\theta, \mathfrak{m}_{\theta^-})$, and is formulated as:

$$\mathcal{L}_e(\mathfrak{m}_\theta, \mathfrak{m}_{\theta^-}) = \frac{1}{\aleph}\sum_{i=1}^{\aleph} \left\| \mathscr{s}\left(\mathfrak{m}_{\theta_i}\right) - \mathscr{s}\left(\mathfrak{m}_{\theta^-_i}\right) \right\|_2^2, \tag{12}$$

where $\mathscr{s}$ is the Sobel operator [84] used to detect the boundary contours of obstacles in $\mathfrak{m}_\theta$, and $\mathfrak{m}_{\theta^-}$. By combining LPIPS, and $\mathcal{L}_e$, $\mathcal{L}_m$ is formulated as:

$$\begin{aligned} \mathcal{L}_m = {} & \mathscr{a} \cdot \mathrm{LPIPS}(\mathfrak{m}_\theta, \mathfrak{m}_{\theta^-}) \\ & + \mathscr{b} \cdot \mathcal{L}_e(\mathfrak{m}_\theta, \mathfrak{m}_{\theta^-}), \end{aligned} \tag{13}$$

where $\mathscr{a}$ and $\mathscr{b}$ are scaling hyperparameters for the LPIPS and $\mathcal{L}_e$ components, respectively. The output of the *MPN* is the predicted robot heightmap $\widehat{M_\iota}$ which is provided to both the *Confidence Network* and *Exploration Planner* modules.

*3) Confidence Network*

The *CN* is used to measure the uncertainty of $\widehat{M_\iota}$, to guide robot exploration towards frontier regions with higher prediction uncertainty. Inputs into the *CN* include a single three channel image, where the first channel contains the predicted map $\widehat{M_\iota}$, the second channel contains the nearby robot trajectory image $\widehat{\delta_t}$, and the third channel contains a binary mask for the observed region, $M_i^{mobs}$.

The *CN* uses a Residual Fully Convolutional Variational Autoencoder (RFC-AEM) model to produce a confidence map, $C_i$, which represents the uncertainty of each predicted cell $m_{(x,y)}$ within $\widehat{M_\iota}$. The RFC-AEM consists of an encoder and decoder, Fig. 2. The encoder includes six CBs. These blocks facilitate map feature extraction and utilize a latent space bottleneck for dimensionality reduction to capture only the most salient features from the input map. The decoder mirrors the encoder with six transposed CBs, enabling the reconstruction of detailed confidence maps from the compact encoded latent representations of the encoder. Herein, we chose to have a separate autoencoder design for confidence prediction as it enables task-specific optimization, where the *CN* is trained specifically to capture epistemic uncertainty in map predictions.

The *CN* adopts a self-supervised training paradigm. Ground truth confidence maps, $C_i^{GT}$, are obtained by comparing the prediction, $\widehat{M_\iota}$, with the ground truth heightmap, $M_h$, where each value in $C_i^{GT}$ is a binary indicator, specifying whether the prediction is correct or incorrect. To train the *CN*, a pixel-wise Mean Squared Error (MSE) loss function is utilized:

$$\mathcal{L}_{\mathrm{MSE}} = \frac{1}{\aleph}\sum_{x,y\in\widehat{M_\iota}} \left(C_i(x,y) - C_i^{GT}(x,y)\right)^2, \tag{14}$$

where $\aleph$ represents the total number of pixels in $\widehat{M_\iota}$, and $x, y$ are the pixel coordinates of $\widehat{M_\iota}$. Minimizing $\mathcal{L}_{\mathrm{MSE}}$, minimizes the uncertainty measure between $\widehat{M_\iota}$ and $M_h$. The predicted confidence map $C_i$ is provided to the *Exploration Planner* for frontier exploration.

*C. Exploration Planner*

The objective of the *Exploration Planner* subsystem is to: 1) choose frontiers that maximize the utility function, $U$, while maintaining the energy budget of a robot, Eq. (6), using the *Frontier Selection* module, and 2) have the robot navigate towards the frontier goal using the *Navigation Controller*. The *Frontier Selection* module, $\partial$, uses the robot's predicted map $\widehat{M_\iota}$ and the associated confidence map $C_i$ to evaluate the utility, $u$, of a frontier location, $g(x,y)$, and selecting the $g$ with the highest $u$. The utility function, $U(g)$, calculates the $u$ of each $g$ based on the expected information gain $I$, traversability score $\mathcal{T}$, and travel distance to goal $D$:

$$U(g) = \alpha \times I + \beta \times \mathcal{T} + \gamma \times D. \tag{15}$$

The expected information gain is determined by the estimated traversable area and its uncertainty. Specifically, $I$ is computed by evaluating the traversable areas, $A_{d_r,}$, in the predicted map, $\widehat{M_\iota}$, within a fixed radius, $d_r$, around the frontier position, $g$. Note, only the predicted region, $M_i^{pred}$, within $\widehat{M_\iota}$ is considered. This evaluation involves integrating the traversable area values in $\widehat{M}_l$ which are weighted by the corresponding uncertainty scores from $C_i$ over $A_{d_r,}$. $I$ is then obtained by averaging $\int_{A_{d_r,}} \widehat{M}_l \cdot C_i \, dA$ over the total area of $A_{d_r}$:

$$I = \frac{1}{\left|A_{d_r,}\right|}\int_{A_{d_r,}} \widehat{M}_l \cdot C_i \, dA. \tag{16}$$

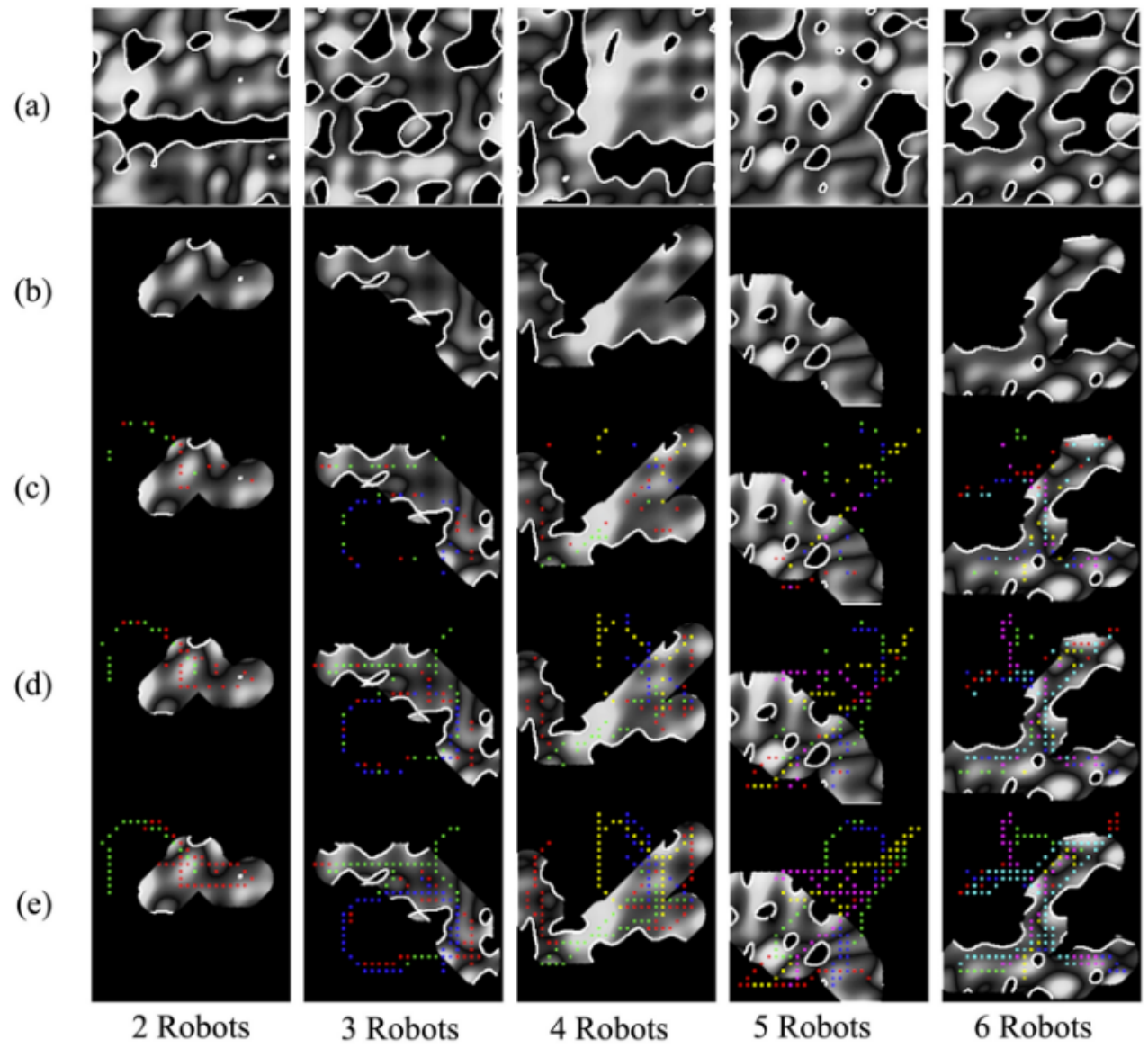

**Fig. 4.** Robot heightmaps for scenes with 2 to 6 robots: (a) Ground truth map of environment with irregularly shaped obstacles (white lines) and uneven terrain (grayscale gradient for terrain height); and (b) Input map of observed area, unobserved (masked) regions are black. The dotted color lines represent communicated trajectories from different robots at: (c) 25% CSP, (d) 50% CSP, and (e) 100% CSP.

The traversability score $\mathcal{T}$ is the summation of elevation values $m_{(x,y)}$ along the shortest collision-free path between the robot's current position $p$ and the frontier $g$ denoted as $\tau_{p\rightarrow g}$. Thus, the traversability score is described by:

$$\mathcal{T} = \sum_{\tau_{p\rightarrow g}} m_{(x,y)}. \quad (17)$$

$D$ is calculated as the length of $\tau_{p\rightarrow g}$. The coefficients $[\alpha, \beta, \gamma]$ in Eq. (15) are determined through domain expert tuning in order to prioritize frontiers that maximize $I$ while minimizing $\mathcal{T}$ and $D$. The frontier $g$ with the highest $u$ is used by the *Navigation Controller* module for global path planning using A*. Local path planning is achieved with Timed Elastic Band Planner [85], which adjusts the robot's trajectory based on real-time sensor data to avoid both static and dynamic obstacles.

## V. Datasets

We developed two simulated datasets to train *4CNet.* These datasets include: 1) a map-trajectory dataset for training the *Trajectory Encoder* and *MPN* modules, and 2) a prediction map dataset used for the training of the *CN*.

**Map-Trajectory Dataset, $\mathcal{D}_{\mathrm{MT}}$:** 2D heightmaps, $M_h$, were generated using the Diamond-Square Algorithm [86], Fig. 4(a), with irregularly shaped obstacles (white lines), and uneven terrain (grayscale gradients). These irregularly shaped obstacles represent natural outdoor elements such as flower beds, tree clusters, or bushes, while the uneven terrain simulates natural variations like small hills. Each heightmap has a resolution of $224 \times 224$ pixels which represents a spatial area of $15 \times 15$ m. The number of robots in each environment is randomly varied from 2 to 6 robots, with each robot having random feasible start and end positions in the environment. The A* search algorithm [72] was used to generate collision-free paths between these positions for each heightmap. Since *4CNet* processes trajectories as a Binary Trajectory Image, $\widehat{\delta_t}$, the trajectories of all robots are aggregated into a single uniform trajectory, rather than distinct, overlapping trajectories. To simulate communication dropout, CSP was applied to probabilistically remove robot positions from the A* generated trajectories at dropout rates of 75%, 50%, and 0% (Eq. (4)), Fig. 4(c)-(e). Namely, as CSP increases, less positions within the exchanged trajectories are dropped. On the other hand, increasing team size results in increasing trajectory length. Varying team size in the dataset is important to train the trajectory encoder to generate embeddings that are invariant to the trajectory length. In total, the dataset includes 75,000 pairs of heightmaps, $M_h$, and robot trajectories, $\delta_t$. For evaluation purposes, 20% of the $\mathcal{D}_{\mathrm{MT}}$ was used as the test set, $\mathcal{D}_{\mathrm{MT-T}}$.

**Predicted Map Dataset, $\mathcal{D}_{\mathrm{PM}}$:** The Predicted Map Dataset contains 20,000 samples of: 1) 2D predicted heightmaps, $\widehat{M_\iota}$, 2) ground truth heightmaps, $M_h$, and 3) corresponding robot trajectories, $\delta_i$. The predicted heightmaps were generated by randomly masking regions of the ground truth heightmaps in the $\mathcal{D}_{\mathrm{MT}}$ to simulate partial maps (Fig. 4(b)) and using the *MPN* to generate the predicted map $\widehat{M_\iota}$. Each $\widehat{M_\iota}$ is paired with its $M_h$ as well as the corresponding $\delta_i$ from the $\mathcal{D}_{\mathrm{MT}}$. For evaluation purposes, 20% of the $\mathcal{D}_{\mathrm{PM}}$ was used as the test set, $\mathcal{D}_{\mathrm{PM-T}}$.

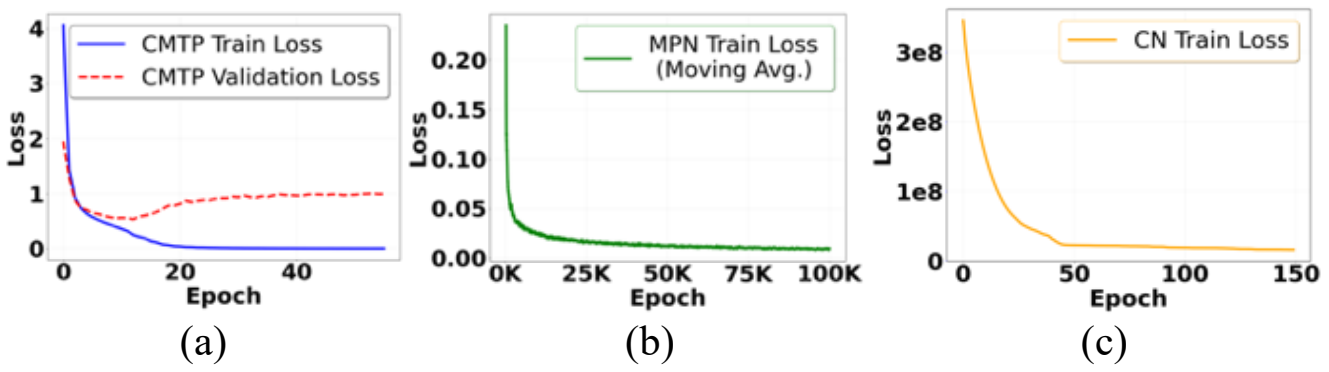

**Fig. 5.** (a) *CMTP* training and validation losses, (b) *MPN* training loss, and (c) *CN* training loss.

## VI. Training of 4CNet

Training consisted of: 1) using the proposed *CMTP* framework to train the *Trajectory Encoder*, 2) training of the *Map Prediction Network* conditioned on robot trajectory embeddings, and 3) training of the *Confidence Network* based on the predictions of the *MPN*. All training was conducted on a workstation with i9-13900KF Intel CPU, Nvidia RTX 4090 GPU, and 64 GB of RAM.

### A. Contrastive Map Trajectory Pretraining

The *CMTP* framework was used to train the Trajectory and Map Encoder modules in parallel, as described in Section IV.B.1. Namely, *CMTP* was trained with a batch size of 128. A neural dropout rate of 0.3 was used to promote generalizability of the model and minimize overfitting [87]. We used a learning rate of 0.001 for gradient descent optimization, and a temperature of 1 was selected for the softmax function in the contrastive loss, Eq. (7). The *CMTP* was trained for 10 hours over a duration of 56 epochs, Fig. 5(a). We implemented early stopping and obtained the lowest validation loss at the 15th epoch (approximately 2.5 hours of training).

### B. Map Prediction Network Training

The *Map Prediction Network* was trained with a batch size of 8. The hyperparameters of the minimum and maximum standard deviations for the PF ODE noise were set to 0.002 and 80, similar to [78]. A Keras schedule hyperparameter of 7 was used with initial and final PF ODE time steps of 2 and 100. The initial exponential moving average decay rate of 0.95 was used for the target network $\boldsymbol{f}_{\boldsymbol{\theta}^-}$, and a learning rate of 0.00002 was utilized during training [78]. The scaling factors $a$ and $b$ from Eq. (13) were set to 0.1 and 0.04, respectively. The *MPN* was trained for a total of 100,000 steps over 37 hours. The training loss graph is presented in Fig. 5(b). The loss converged to 0.01 by 100,000 training epochs.

### *C. Confidence Network Training*

The *Confidence Network* was trained with a batch size of 64 and a learning rate of 0.0001. The training loss converged to $2.0 \times 10^6$ within 100 epochs after 4.5 hours of training, Fig. 5(c).

## VII. Simulated Experiments

We conducted four simulated experiments to evaluate the performance of *4CNet* and *4CNet-E* using: 1) a comparison study to evaluate the map prediction performance of *4CNet* against heuristic and DL methods, 2) a comparison study of robot coverage during resource-limited exploration between *4CNet-E* and existing exploration methods, 3) an ablation study to investigate the design choices of *4CNet*, and 4) a runtime analysis to evaluate the computational cost of *4CNet*. We set the total number of denoising time steps, $t_{total}$, as 30, to enable multi-pass prediction using *4CNet*; thereby, allowing iterative refinement of the generated map during robot map prediction.

### *A. Comparison Study for Map Prediction Performance*

A comparison study was performed to evaluate our *4CNet* subsystem against state-of-the-art (SOTA) heuristic and learning methods for different number of robots and CSPs. We measured map prediction performance using the following metrics: 1) MSE for overall prediction error; 2) Obstacle Intersection over Union (OIOU) to measure only the ratio of overlapping obstacle pixels between the predicted map and the ground truth map in order to evaluate the accuracy of predicted obstacle locations and shapes; 3) Feature Similarity Index (FSIM) [88] to assess the structural and feature similarity of predicted and ground truth heightmaps; and 4) Valid Trajectory Score (VTS), a metric we created to measure the proportion of a robot's trajectory that is on traversable terrain relative to the entire trajectory in the predicted heightmap: $\text{VTS} = \tau_i^v/\tau_i$. $\tau_i^v$ denotes the sequence of robot positions within a robot trajectory that coincides with traversable terrain (non-obstacle space).

*1) Comparison Methods:* We compared with three methods.

**1. Database-based method (DB)** [20]**:** The DB method is a heuristic-based approach which selects a reference heightmap from a database based on feature resemblance. The reference map is then integrated into the unexplored target area using Gaussian filtering to create a predicted robot heightmap, Fig. 6(a).

**2. Autoencoder Model method (AEM)** [14]**:** The AEM method is a DL approach consisting of encoder and decoder networks, Fig. 6(b). Both networks utilize a ResNet50 architecture with skip connections [14]. The input to the AEM is a two-channel image: the first channel contains the grayscale partially observed robot heightmap, and the second channel consists of the mask of the target region. The mask has dimensions of $80 \times 80$ pixels, the same as in [14]. The encoder network performs down sampling of the partially complete robot heightmap to capture spatial features. The decoder network reconstructs these spatial features from latent variables obtained from the encoder network, producing a predicted map. To achieve prediction of the entire unknown region, a sliding window technique is implemented. Predictions are then made in a cascading manner.

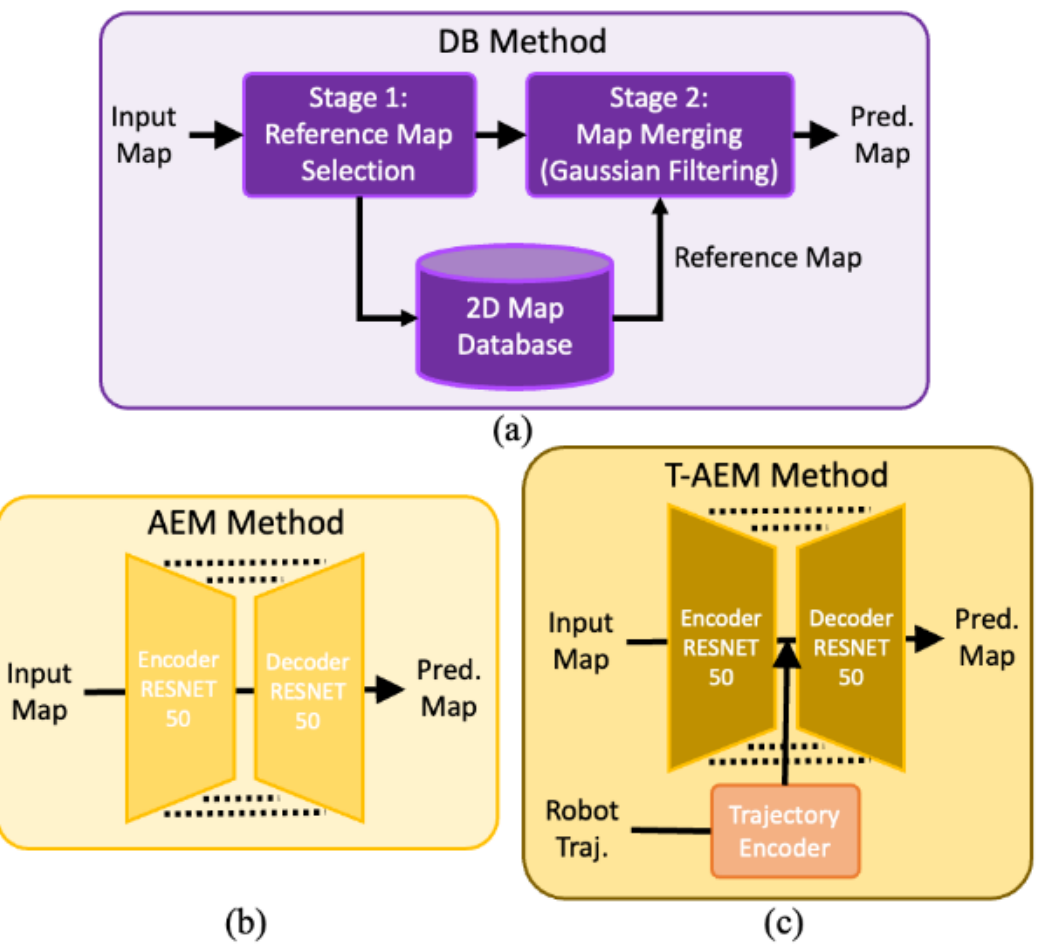

**Fig. 6.** (a) Database-based (DB) method, (b) Autoencoder model (AEM) method, and (c) Trajectory conditioned AEM method (T-AEM).

**3. Trajectory Conditioned AEM method (T-AEM):** We extended the AEM method to incorporate robot trajectory data for map prediction, i.e., T-AEM, as to the authors' knowledge, trajectory conditioned map prediction methods do not currently exist. In T-AEM, the inputs are the same as the AEM approach with the addition of robot trajectory embeddings within the latent space, Fig. 6(c). Namely, we utilized the *Trajectory Encoder* module from our proposed *4CNet* model to allow T-AEM to condition its map predictions on nearby robot trajectories, similar to *4CNet*. The T-AEM is used as a benchmark to evaluate the impact of adding trajectory information to a current SOTA method.

For the DB method, we randomly selected 5,000 heightmaps from the $\mathcal{D}_{\text{MT}}$ database (the same number of maps as in [20]) during each prediction, to balance reference map quality and search speed. Both AEM and T-AEM were trained on $\mathcal{D}_{\text{MT}}$.

*2) Comparison Results:* Table I and Figure 7 provide a comparative analysis of our *4CNet* and the SOTA map prediction methods. Each prediction method used as input a masked heightmap from $\mathcal{D}_{\text{MT-T}}$, i.e., Fig. 7(a). In general, our *4CNet* achieved the lowest average MSE and highest average OIOU, FSIM and VTS at each CSP level across the number of robots. The DB method had the lowest performance among the methods. This was due to its reliance on a priori reference maps in its dataset to accurately represent an unknown environment. Therefore, when new obstacles that were not represented in the database were observed, the map prediction from DB resulted in misaligned and disjointed obstacle prediction. Furthermore, the use of Gaussian blurring to integrate these misaligned maps

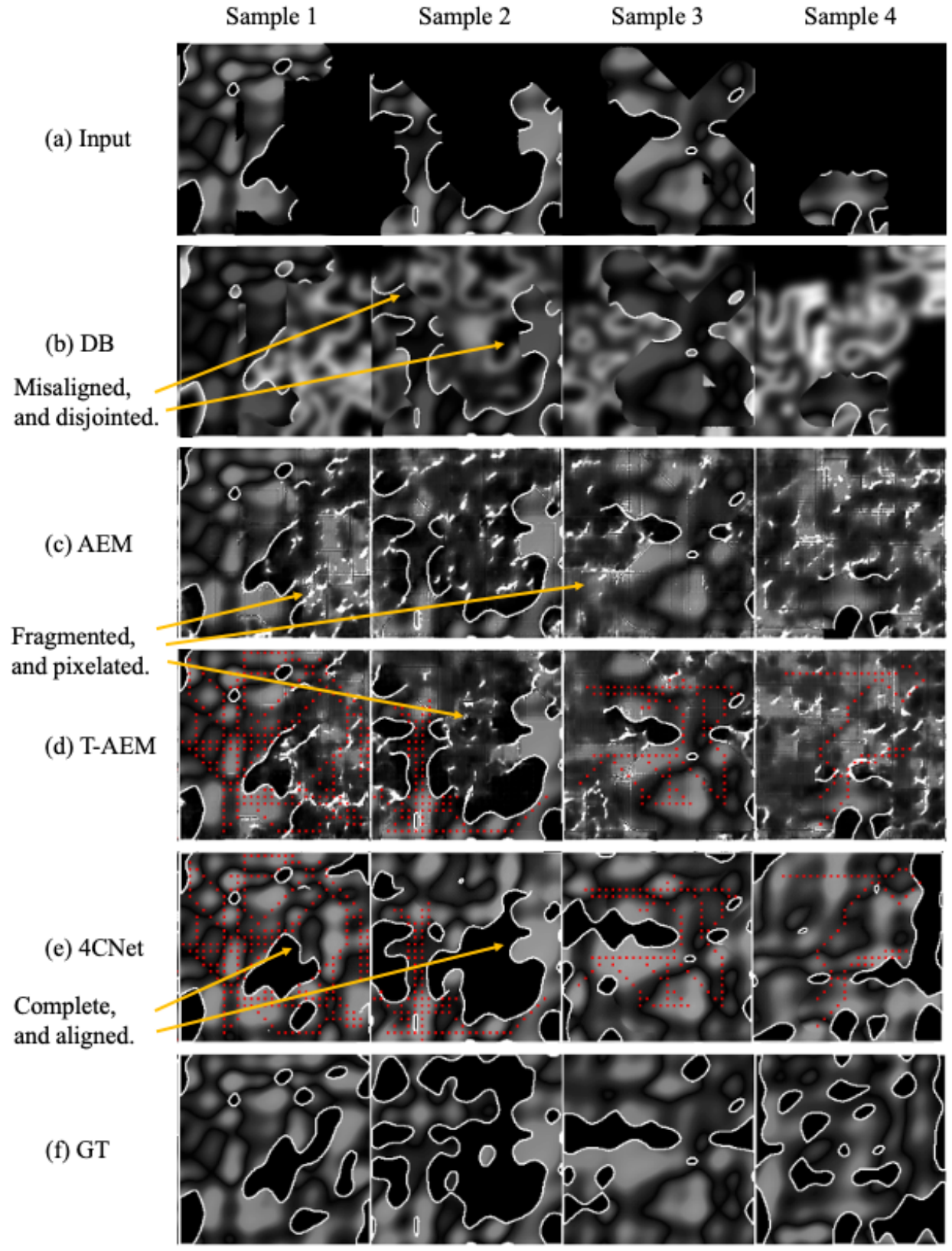

**Fig. 7.** Predicted map results using the: (a) partially completed maps from $\mathcal{D}_{\mathbf{MT-T}}$ as input for the: (b) DB, (c) AEM, (d) T-AEM, and (e) *4CNet* methods, compared to (f) ground truth (GT) maps. Red dotted lines denote communicated robot trajectories.

with actual observed maps resulted in an overall blurred predicted map, as shown in Fig. 7(b).

The AEM method also had higher MSE and lower OIOU, FSIM and VTS than *4CNet*. In particular, the lower OIOU and VTS of 0.24 and 0.46 compared to *4CNet* (0.59, and 0.91 with 100% CSP) was due to AEM not being able to accurately reconstruct obstacles, leading to fragmented and pixelated predictions, Fig. 7(c). This fragmentation was primarily due to the single-pass prediction of AEM.

As the T-AEM method incorporated robot trajectories during map prediction, it had a statistically significant improvement as defined by Friedman tests ($p < 0.001$), in terms of MSE, OIOU and FSIM when compared to AEM with 100% CSP. This showed robot trajectories improved heightmap prediction. However, it is interesting to note that since T-AEM had similar VTS to AEM, the robot trajectory embeddings did not seem to improve the obstacle prediction accuracy of T-AEM. This is due to only a portion of the communicated robot trajectory being considered within each fixed dimension prediction window, leading to the incomplete and fragmented obstacle predictions by T-AEM, Fig. 7(d).

The better performance of *4CNet* in both single- and multi-robot scenarios was primarily due to the advantages of consistency models in: 1) multi-pass prediction, and 2) the use of arbitrarily shaped and sized target regions during prediction, which were not limited to a fixed dimension. Namely, the multi-pass prediction allowed for refinement of the map over a sequence of denoising time steps, resulting in predictions, Fig. 7(e), that closely matched the ground truth maps, Fig. 7(f).

TABLE I
MAP PREDICTION PERFORMANCE COMPARISON

| Method | CSP | # of Robots | MSE$^{\downarrow}$ | OIOU$^{\uparrow}$ | FSIM$^{\uparrow}$ | VTS$^{\uparrow}$ |
|---|---|---|---|---|---|---|
| **DB** | - | 1 | 78.47 | 0.29 | 0.36 | 0.15 |
| **AEM** | - | 1 | 69.18 | 0.24 | 0.41 | 0.46 |
| **T-AEM** | 25% | 2-6 | 74.53 | 0.22 | 0.40 | 0.45 |
| | 50% | 2-6 | 67.37 | 0.26 | 0.43 | 0.48 |
| | 100% | 2-6 | 64.97 | 0.28 | 0.44 | 0.47 |
| **4CNet** (ours) | - | 1 | 56.11 | 0.51 | 0.51 | 0.79 |
| | 25% | 2-6 | 56.23 | 0.52 | 0.49 | 0.82 |
| | 50% | 2-6 | 53.20 | 0.58 | 0.53 | 0.89 |
| | 100% | 2-6 | 50.35 | 0.59 | 0.55 | 0.91 |

↑ indicates a higher value represents better performance.
↓ indicates a lower value represents better performance.

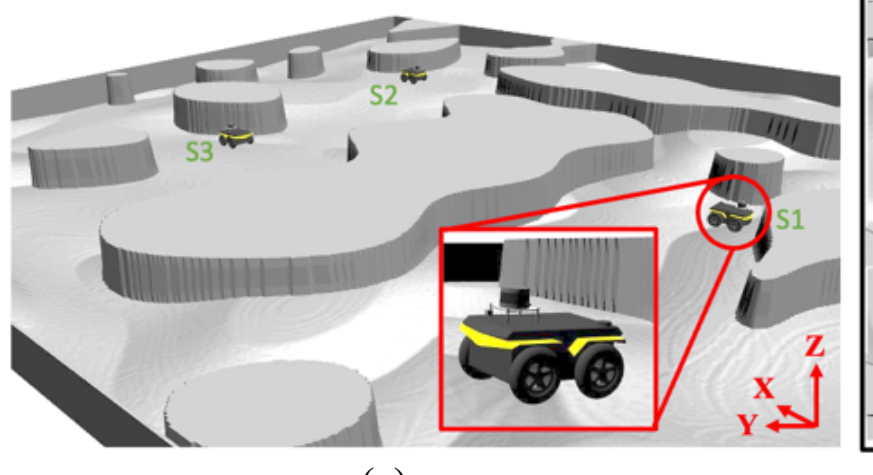

(a)

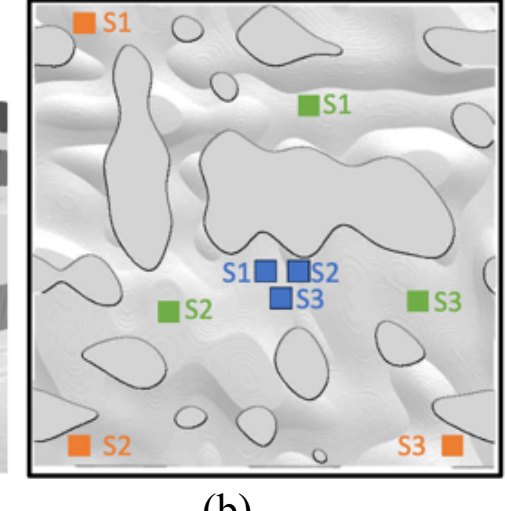

(b)

**Fig. 8.** (a) Robots in 3D simulation environment with irregularly shaped obstacles and uneven terrain. (b) Three sets of initial positions for each of the 3 mobile robots. S1-3 denote the starting positions of each robot.

This was further supported by *4CNet* having the lowest MSE, and highest FSIM. As *4CNet* was able to account for varying target regions and thus, effectively considered the entire robot trajectories communicated during each prediction. Furthermore, *4CNet* having the highest OIOU and VTS was due to its ability to better predict obstacles in the environment, Fig. 7(e).

With respect to CSPs, *4CNet* achieved lower MSE, and higher OIOU, FSIM, and VTS compared to T-AEM across all CSPs. Friedman Tests conducted across the MSE, OIOU, FSIM and VTS metrics for the AEM and DB methods and all CSPs for the *4CNet* and T-AEM methods found statistically significant differences existed ($p < 0.001$). Post-hoc Wilcoxon Signed-rank tests with a Bonferroni correction confirmed a statistically significant difference in all four metrics when *4CNet* was compared with each SOTA method, across all CSPs ($p < 0.0167$). Therefore, validating *4CNet*'s improved performance over the other methods.

### B. *Comparison Study for Exploration Performance*

We evaluated the performance of *4CNet-E* in 3D simulated resource-limited environments with irregularly shaped obstacles and uneven terrain. We introduced two environment sizes and three energy budgets. Four robot exploration methods were compared with *4CNet-E*: namely, exploration methods with 1) no map prediction, 2) AEM map prediction, 3) T-AEM map prediction, and 4) 4CNet without CN map prediction. We measured the percentage of area coverage for each energy budget.

*1) Mobile Robots:* Three Clearpath Jackal mobile robots were used, Fig. 8(a). Each robot was equipped with a 360-degree LiDAR with a sensing range of 1.5 m, and both wheel encoders and an inertial measurement unit (IMU). Communication of robot trajectories was facilitated when the robots were within the aforementioned sensing range using 100% CSP.

*2) Frontier Selection*: Frontiers were selected using the utility equation in Eq. (15). The coefficients were set to [4, -1, -5] based on an expert-guided search, with the aim to maximize $I$ while minimize $\mathcal{T}$ and $D$.

*3) Environment:* Two 3D environments were randomly generated using ROS Gazebo, consisting of uneven traversable terrain and irregularly shaped non-traversable obstacles, Fig. 8(a). The sizes of these environments were $15\ \mathrm{m} \times 15\ \mathrm{m}$ ($225\ \mathrm{m}^2$) and $30\ \mathrm{m} \times 30\ \mathrm{m}$ ($900\ \mathrm{m}^2$). The elevation of the traversable terrain ranged from 0 to 0.3 m to represent uneven slopes for the Clearpath Jackal robots to traverse. Obstacle heights were 0.7 m in the *z*-axis and were not traversable. Three sets of initial robot positions were used as shown in Fig. 8(b): where robots started 1) at opposite ends of the environment, 2) in the center nearby each other, and 3) random locations with at least a minimum distance of 3 meters.

*4) Energy Budgets:* Robots operated under three distinct energy budget levels: Low, Medium, and High. These energy budgets allowed the robots to only explore a portion of an environment, promoting the use of map prediction in the exploration process. The initial energy budgets provided to the robots were empirically determined to cover 20%, 40% and 60% of the environment with *4CNet-E*. These exploration percentages were chosen to evaluate the prediction performance of *4CNet-E* under three increasing levels of exploration coverage. For all experiments, we have set $w_i = 1$, $k_i = 2$ for Equation 1 and $\alpha = \beta = \gamma = 0.33$ for Equation 15.

*5) Comparison Methods:* We compared our *4CNet* exploration method (*4CNet-E*) against both non-predictive and predictive exploration methods. All exploration methods utilized the same frontier selection approach, Eq. (15), in order to directly compare performance in terms of map prediction influence on exploration. Furthermore, the comparison methods utilized a uniform uncertainty score in estimating the expected information gain, as they do not have a method for predicting the map prediction uncertainty.

**1. Non-Predictive Exploration (NPE):** NPE is a non-predictive frontier-based exploration method where information gain is naively estimated by assuming all unobserved areas within a frontier region are traversable. NPE is selected to investigate the influence of map prediction on exploration of unknown environments.

**2. AEM Exploration (AEM-E)** [14]: AEM-E utilizes the SOTA AEM model for map prediction for frontier-based exploration.

**3. T-AEM Exploration (T-AEM-E):** T-AEM-E integrates the T-AEM map prediction approach for frontier-based exploration.

**4. 4CNet without CN Exploration (4CNet-C-E):** *4CNet-C-E* is a predictive exploration method that utilizes *4CNet* for map prediction; however, without the *CN* module to predict uncertainty scores. This method is used to investigate the impact of uncertainty scores in the evaluation of expected information gain during frontier selection for *4CNet*.

**5. 4CNet-E-UB:** *4CNet-E-UB* represents the upper bound performance of *4CNet-E*. It utilizes the *4CNet-E* framework; however, it assumes that the robots have perfect map prediction accuracy. Each robot is provided the ground truth map to the robot at every prediction cycle.

*6) Procedure:* A total of 108 trials were conducted in the two environment sizes, using the three energy budgets. Each combination of environment size and energy budget was repeated three times with the different initial robot positions in Fig. 8(b).

*7) Results:* Table II presents the percentage of area coverage with respect to the three energy budgets for *4CNet-E* and the comparison exploration methods in both environment sizes. In general, *4CNet-E* achieved the highest percentage of area coverage regardless of environment size and energy budget.

NPE achieved a lower coverage percentage than *4CNet-E* as it treated unobserved areas as traversable. This resulted in inaccurate information gain estimations, as unobserved regions contained both free space and obstacles. AEM-E and T-AEM-E both utilized predicted maps for expected information gain estimation, however, their lower coverage compared to *4CNet-E* was mainly due to inaccuracies in their map predictions as a result of the aforementioned single pass prediction and fixed target prediction window.

*4CNet-C-E* had higher coverage than NPE, AEM-E and T-AEM-E. However, since *4CNet-C-E* assumes all map predictions have equal confidence (uniform uncertainty score), it had lower area coverage compared to *4CNet-E*. *4CNet-E*, on the other hand, uses the predicted uncertainty scores to weight pixels in the predicted map; thereby, improving accuracy of expected information gain estimates for each exploration frontier. As a result, *4CNet-E* enabled each robot to prioritize exploration in areas of high uncertainty (low confidence), and directly obtaining observations in these regions, to reduce uncertainty in future map predictions, leading to more information gain through coverage. *4CNet-E* achieves the closest exploration coverage to *4CNet-E-UB,* and remains within 4% to 15% of the upper bound, across the two environments and varying energy budgets. This demonstrates that higher map prediction accuracy results in higher exploration coverage. The Friedman Test showed a statistically significant difference was found for percentage of area coverage across the environment sizes and energy budgets for all exploration methods ($p < 0.001$). Post-hoc Wilcoxon Signed-rank tests with a Bonferroni correction between *4CNet-E* and each exploration method showed that *4CNet-E* had a statistically significant higher area coverage than each of these exploration methods ($p < 0.0125$).

TABLE II
COMPARISON OF EXPLORATION AREA COVERAGE (PERCENTAGE)

| Env. Size | $15\ \mathrm{m} \times 15\ \mathrm{m}$ ($225\ \mathrm{m}^2$) | | | $30\ \mathrm{m} \times 30\ \mathrm{m}$ ($900\ \mathrm{m}^2$) | | |
|---|---|---|---|---|---|---|
| Energy / Method | Low | Med | High | Low | Med | High |
| NPE | 18% | 30% | 39% | 18% | 31% | 42% |
| AEM-E | 18% | 30% | 40% | 18% | 30% | 42% |
| T-AEM-E | 18% | 31% | 42% | 19% | 32% | 43% |
| 4CNet-C-E (ours) | 21% | 36% | 49% | 18% | 34% | 48% |
| 4CNet-E (our method) | 23% | 41% | 56% | 21% | 42% | 58% |
| 4CNet-E-UB (upper bound) | 27% | 54% | 71% | 23% | 46% | 68% |

### C. Ablation Study

We performed two ablation studies to evaluate individual design components of *4CNet*. The first study evaluated the impact of the *Trajectory Encoder*, cross-attention mechanism and the *MPN*'s architecture on map prediction accuracy. The second study focused on the effect of the *CN*'s architecture on confidence prediction accuracy.

*1) Map Prediction Ablation Study:* Four variants of *4CNet* were evaluated: 1) *4CNet* with 2 CBs in MPN, and 2) *4CNet* with 4 CBs in MPN, to evaluate the impact of MPN's network depth on map prediction accuracy. 3) *4CNet* w/o Trajectory Encoder: This configuration removed the trajectory embedding vector, which provides spatial information from nearby robots, to evaluate its influence on prediction accuracy. 4) *4CNet* w/o Cross-Attention: We replaced the cross-attention mechanism in the *MPN* with a direct concatenation of the trajectory embeddings to the latent space to assess the importance of selective attention during the prediction process. The performance metrics followed Section VII.A, which included MSE, OIOU, FSIM, and VTS to evaluate the accuracy of the predicted map with respect to the ground truth map. We use $\mathcal{D}_{\mathrm{MT-T}}$ for this ablation study.

The map prediction ablation results are shown in Table III. The *4CNet* with 2 CBs achieved the worst performance across all metrics, with an MSE of 88.21, OIOU of 0.21, FSIM of 0.32, and VTS of 0.12. This result indicated that the shallow network depth negatively impacted the *MPN*'s ability to capture spatial features which resulted in poor map prediction accuracy. Increasing the depth from 2 to 4 CBs showed improved map prediction results, achieving an MSE of 54.32, OIOU of 0.52, FSIM of 0.52, and VTS of 0.87. However, *4CNet* with 4 CBs still underperformed compared to our complete *4CNet* architecture with 5 CBs. The *4CNet* w/o the Trajectory Encoder, which excluded spatial information from nearby robots, also showed a decline in performance compared to *4CNet*, with an MSE of 57.93, OIOU of 0.51, FSIM of 0.51, and VTS of 0.80. This result was similar to the single-robot scenario (as shown in Table I), showing that the absence of multi-robot trajectory information degraded the *MPN*'s ability to predict the map accurately. Lastly, the *4CNet* w/o the Cross-Attention mechanism achieved the best performance among all ablation variants, with an MSE of 53.21, OIOU of 0.52, FSIM of 0.54, and VTS of 0.84. The results indicated that while the FSIM value without the cross-attention mechanism (0.53) was similar to the full *4CNet* model (0.54), the MSE, OIOU, and VTS were lower. This showed that cross-attention was necessary to capture long-term dependencies between robot trajectories for accurate obstacle contour prediction.

TABLE III
MAP PREDICTION ABLATION RESULT

| Variant | MSE↓ | OIOU↑ | FSIM↑ | VTS↑ |
|---|---|---|---|---|
| **4CNet with 2 CBs in MPN** | 88.21 | 0.21 | 0.32 | 0.12 |
| **4CNet with 4 CBs in MPN** | 54.32 | 0.52 | 0.52 | 0.87 |
| **4CNet w/o Trajectory Encoder** | 57.93 | 0.51 | 0.51 | 0.80 |
| **4CNet w/o Cross-Attention** | 53.21 | 0.52 | 0.53 | 0.84 |
| **4CNet (ours)** | 51.47 | 0.57 | 0.54 | 0.92 |

*2) Confidence Prediction Ablation Study:* For the confidence prediction ablation study, we examined two variants of the *CN*: 1) CN with 2 CBs, and 2) CN with 4 CBs, to investigate the impact of *CN*'s network depth on confidence prediction accuracy. The performance metric included MSE to evaluate the accuracy of the predicted uncertainty score with respect to the ground truth difference between the predicted map and the ground truth map. The lower the predicted uncertainty score (i.e., pixel error), the higher the prediction confidence. We use $\mathcal{D}_{\mathrm{PM-T}}$ for this ablation study.

The results are shown in Table IV, which showed the effect of varying the number of CBs on confidence prediction accuracy. Specifically, increasing the number of CBs from 2 to 4 resulted in a significant improvement, reducing the MSE from 285.49 to 257.56. As the number of CBs increases, the *CN* can capture spatial features that are important for accurately modeling the uncertainty associated with confidence predictions. Our proposed *CN* architecture with 6 CBs achieved the lowest MSE of 251.21.

TABLE IV
CONFIDENCE PREDICTION ABLATION RESULT

| Variant | MSE↓ |
|---|---|
| **CN w/ 2 CBs** | 285.49 |
| **CN w/ 4 CBs** | 257.56 |
| **CN (ours)** | 251.21 |

### D. Runtime Analysis

We conducted a comprehensive runtime analysis of the entire *4CNet* pipeline to investigate the computational cost of each module within *4CNet* in comparison to the DB and AEM methods. To do this, we measured the average runtime in seconds for each stage of map prediction over 100 map predictions using an NVIDIA RTX 4090 GPU.

The computational cost results are shown in Table V. *4CNet* exhibited an end-to-end computation time of 5.452 seconds. This was longer than AEM, which required 0.0687 seconds, but significantly shorter than DB, which took 85.05 seconds. The DB method's high computation cost was attributed to the extended search process for a reference map within its onboard database. In contrast, AEM achieved the fastest prediction time due to its single pass through its network architecture. *4CNet*'s prediction time showed that the *Trajectory Encoder* required 0.0728 seconds, the *MPN* took 5.319 seconds, and the CN required 0.0598 seconds. The extended runtime of *4CNet* was primarily because of the multi-pass prediction process in the *MPN*, as well as the additional processing to generate a confidence map required by the *CN*.

While the computational cost of *4CNet* is higher than AEM, this trade-off yields two key benefits. First, *4CNet*'s multi-pass approach resulted in more accurate map predictions compared to the other methods, as shown in Table I. Second, the *CN* enabled *4CNet* to generate a confidence map. By leveraging both the accuracy of the predicted map and the confidence map, *4CNet* achieved higher exploration coverage compared to the other methods using the same energy budget, Table II.

TABLE V
COMPUTATION COST ANALYSIS

| Method | Modules | Prediction Time (s) |
|---|---|---|
| **DB** | End-to-end | 85.05 |
| **AEM** | End-to-end | 0.0687 |
| **4CNet** | Trajectory Encoder | 0.0728 |
| | Map Prediction Network | 5.319 |
| | Confidence Network | 0.0598 |
| | End-to-end | 5.452 |

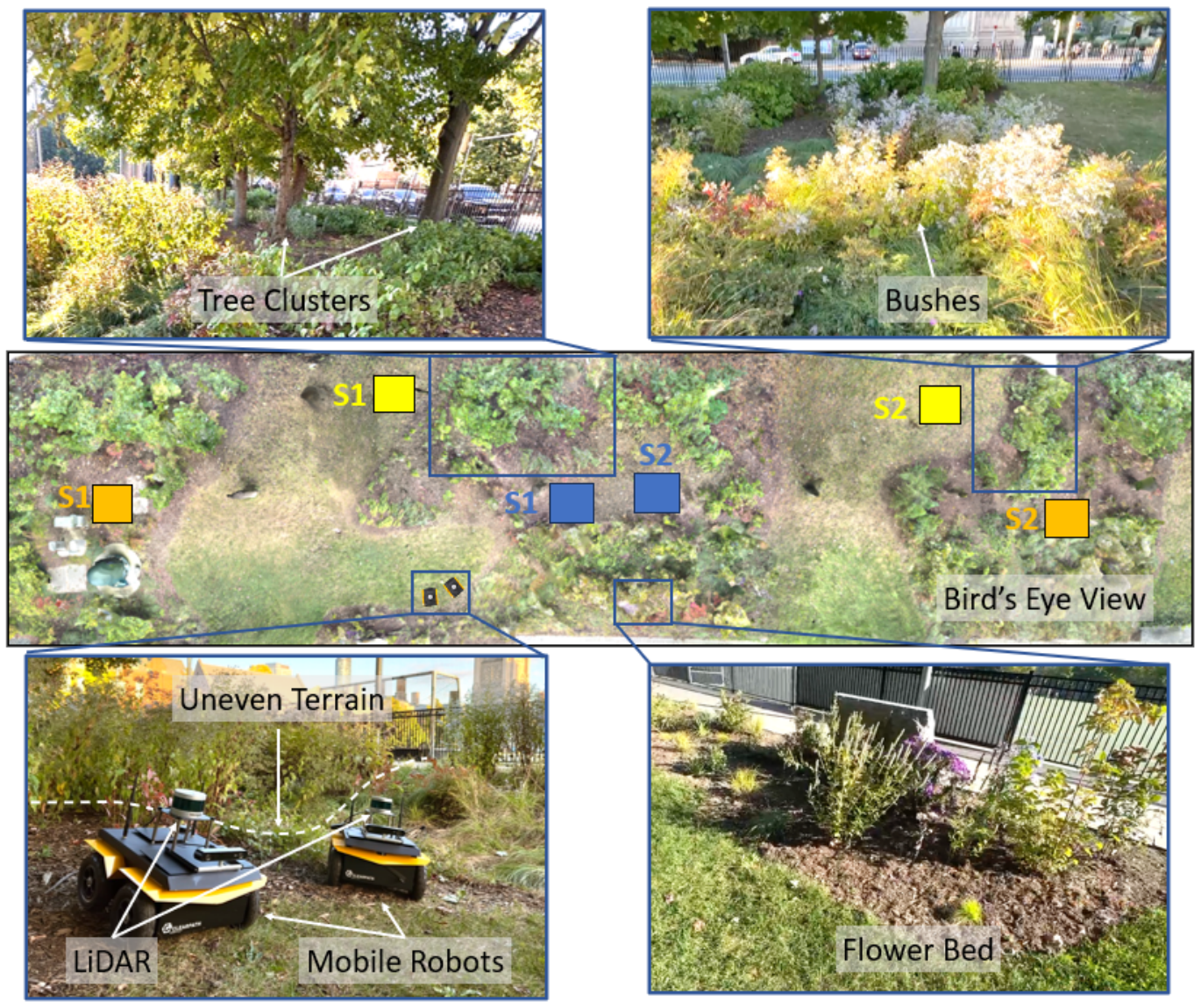

**Fig. 9.** Bird's eye view of a real natural outdoor environment of size 45 × 15 $m$, with irregularly shaped obstacles (tree clusters, bushes and flower beds), and uneven terrain (natural surfaces). Two Clearpath Jackal robots (S1 and S2) equipped with Velodyne 360-degree LiDAR are deployed in three initial positions in the environment; 1) Opposite Ends (Orange): starting at opposite ends, 2) Center (Blue): starting in the center, and 3) Random (Yellow): starting at randomized locations.

## VIII. EXPERIMENTS

We conducted two robot exploration experiments with *4CNet-E* to investigate its performance in two real-world environments. The first experiment was a comparison study with SOTA methods to evaluate the applicability and generalizability of *4CNet-E* with existing methods. The experiment was performed in a real natural outdoor environment of size 45 × 15 m, with irregularly shaped obstacles and uneven terrain. The second experiment was a case study which detailed the step-by-step progress of multi-robot exploration using map and confidence prediction in an unstructured indoor environment of size 8.5 × 8.5 m, containing irregularly shaped obstacles. In both experiments, ground truth heightmaps were generated using RTAB-Map.

### *A. Comparison Experiment of Map Prediction in a Real Natural Outdoor Environment*

In this experiment, we used two mobile Jackal robots from Clearpath Robotics with a 50% CSP for decentralized exploration in a real natural outdoor environment, Fig. 9. Both robots used an embedded Velodyne 360-degree LiDAR (max range 2.5 m) for mapping and map prediction, and are equipped with a NVIDIA GTX 1050 GPU for running map prediction.

An ad-hoc wireless network was used to enable peer-to-peer communication, where the robots exchanged GPS coordinates that present robot trajectories. The communication range was simulated using Received Signal Strength Indicator (RSSI), allowing robots to exchange their trajectories when the signal strength exceeded -50 dBm. The environment consists of irregularly shaped obstacles such as bushes, tree clusters, flower beds and uneven terrain. The ground truth heightmap was obtained with RTAB-Map using a single Jackal robot teleoperated prior to the trial.

Since *4CNet* was trained on an input resolution 224 × 224 pixels in 15 × 15 m environments, we utilize a sliding window and concatenation approach to extend map prediction to larger environments, such as the one shown in Fig. 9. Namely, for the 45 × 15 m environment, we divide the area into four prediction regions. The prediction process begins in the 15 × 15 m region where the robot starts its exploration. To ensure continuity in the output map with the previous predictions, each subsequent input overlaps with approximately 20% of the previously predicted area. This process continues iteratively until the entire environment is predicted.

*1) Experiment Procedure:* We compared *4CNet-E* with AEM-E, and DB-E using the map prediction accuracy metrics of MSE, OIOU, FSIM, and VTS to evaluate the generalizability of our *4CNet* method. *4CNet* was trained on the procedurally generated data ($\boldsymbol{\mathcal{D}}_{\mathbf{MT}}$) discussed in Section V. Three trials were conducted for each method. Map prediction accuracy metrics were averaged for all map predictions in each exploration trial and then averaged across all three trials. The robots traversed the scene at maximum linear and angular velocities of 0.15 m/s and 0.3 rad/s, respectively. Each Jackal robot operated with an energy budget to cover up to 45% of the environment.

*2) Experiment Comparison Results:* The map prediction performance of *4CNet-E* in the outdoor environment is presented in Table VI. Herein, *4CNet-E* outperformed AEM-E and DB-E across all prediction metrics, with lower MSE (63.84), and higher OIOU (0.42), FSIM (0.39) and VTS (0.81). These experimental results are consistent and follow a similar trend with the simulation results presented in Table I. The prediction results are shown in Fig. 10.

TABLE VI
MAP PREDICTION PERFORMANCE COMPARISON

| Methods | MSE$^{\downarrow}$ | OIOU$^{\uparrow}$ | FSIM$^{\uparrow}$ | VTS$^{\uparrow}$ |
|---|---|---|---|---|
| **DB-E** | 88.24 | 0.31 | 0.27 | 0.21 |
| **AEM-E** | 80.65 | 0.16 | 0.29 | 0.23 |
| **4CNet-E (ours)** | 63.84 | 0.42 | 0.39 | 0.81 |

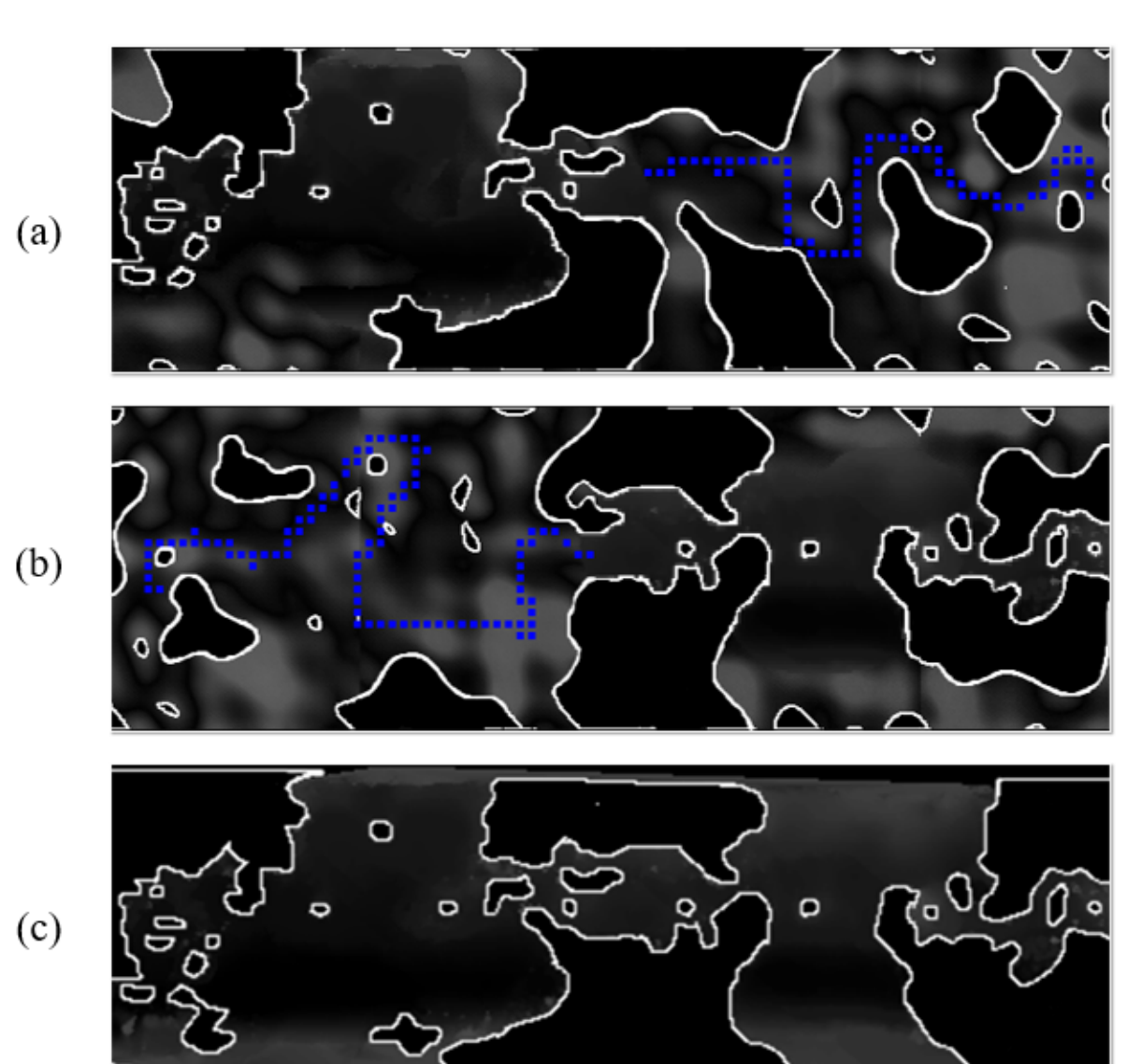

**Fig. 10.** Final predictions for real-world experiments with two robots: (a) Prediction of Robot 1, given Robot 2's trajectory in blue, (b) Prediction of Robot 2 given Robot 1's trajectory in blue, and (c) The ground truth map of the environment.

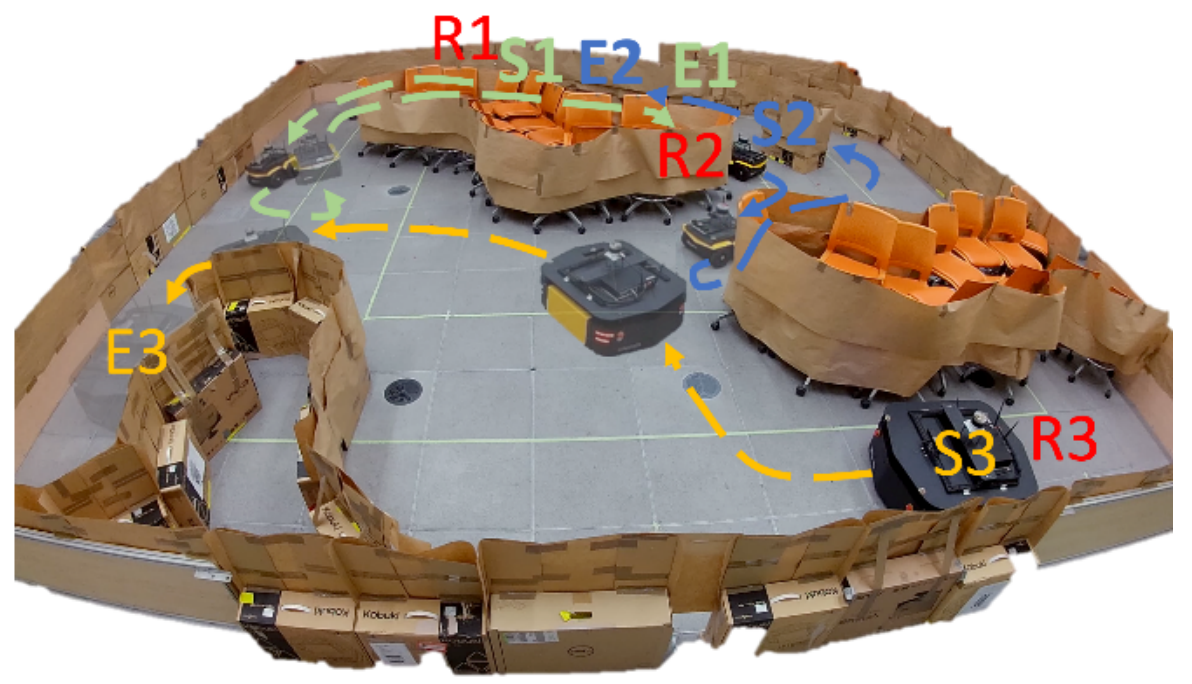

**Fig. 11.** Resource-limited exploration in an unstructured indoor environment with irregularly shaped obstacles and three mobile robots. R1-R3 are Robots 1-3. S1-S3 denote the starting positions of R1-R3. E1-E3 denote the final positions of R1-R3.

*B. Case Study Experiment of Multi-robot Exploration in an Unstructured Indoor Environment*

We conducted an extensive experiment with *4CNet-E* in an 8.5 m by 8.5 m unstructured indoor environment consisting of irregularly shaped obstacles, Fig. 11, with 100% CSP**.** Two Jackal robots and one Ridgeback robot from Clearpath Robotics were deployed with Velodyne 360-degree LiDARs (max range 2.5 m). The ground truth heightmap, and ad-hoc wireless communication between robots were obtained, and set up, respectively, the same as in Section VIII.A. The robots moved with a maximum linear and angular velocities of 0.15 m/s and 0.3 rad/s, respectively. The three robots started at three random start positions, S1, S2 and S3, and ended exploration at end positions E1, E2 and E3, Fig. 11. During the experiment, the two Jackal robots (R1-R2) had an energy budget of 15 m, while the Ridgeback robot (R3) had an energy budget of 8 m. These energy budgets allowed all three robots to achieve up to a maximum of 45% coverage, thereby, requiring map prediction to complete coverage.

*1) Exploration with Map Prediction Results:* Robot map predictions are shown in Fig. 12(a) – (c) at time steps of 0, 103, 196, and 324s. Each robot's observed regions are represented in white and predicted regions are represented in gray. At each time step, individual robots generated map predictions utilizing direct observations $M_i^{obs}$ and trajectory information $\delta_t$ exchanged with nearby robots. In particular, at 103s, R3 and R2 exchanged their trajectories, presented in Figs. 12(b) and (c) by yellow (R3) and blue (R2) dotted lines, respectively. A trajectory exchange also occurred between R3 and R1 at 196s represented by yellow (R3) and green (R1) dotted lines in Fig. 12(b) and (c), respectively. Using both trajectories obtained from R1 and R2, R3 was able to predict the unobserved regions of the obstacle in the middle of the environment in Fig. 12(c). The exploration was completed at 324s, where all three robots depleted their energy budget with an area coverage of 41%, 34% and 35% for R1, R2 and R3, respectively. The obstacle contours of the predicted maps at 324s in Fig. 12(a)-(c), are consistent with the ground truth obstacles, Fig. 12(d), in terms of completeness and alignment.

*2) Confidence Prediction Results:* The generated confidence maps for the three robots are in Fig. 13. Blue and red regions represent low and high uncertainty in the robot map predictions. At 0s, the confidence map for R1 had high uncertainty in the top left region of the environment, Fig. 13(a). As a result, R1 selected a frontier goal in the adjacent area, as indicated by the green Selected Frontier in Fig. 13(a). The confidence map had lower prediction uncertainty in areas with exchanged robot trajectories due to the additional spatial context provided by trajectory embeddings. This was seen at 103s, where the shared trajectories between R3 (yellow dotted line, Fig. 13(b)) and R2 (blue dotted line, Fig. 13(c)) decreased the prediction uncertainties in the corresponding regions as noted by the blue regions. Consequently, at 103s, R2 chose a frontier goal in the upper right area of the environment towards an unexplored region with higher prediction uncertainty, Fig. 13(b). Similar cases of trajectory-influenced confidence and goal selection were observed at 196s and 324s for R1 and R3. A video of our *4CNet-E* approach in both the simulated and real-world unknown environments with irregularly shaped obstacles, and uneven terrain is presented on our YouTube channel at https://youtu.be/nA2a3XXL5Dg.

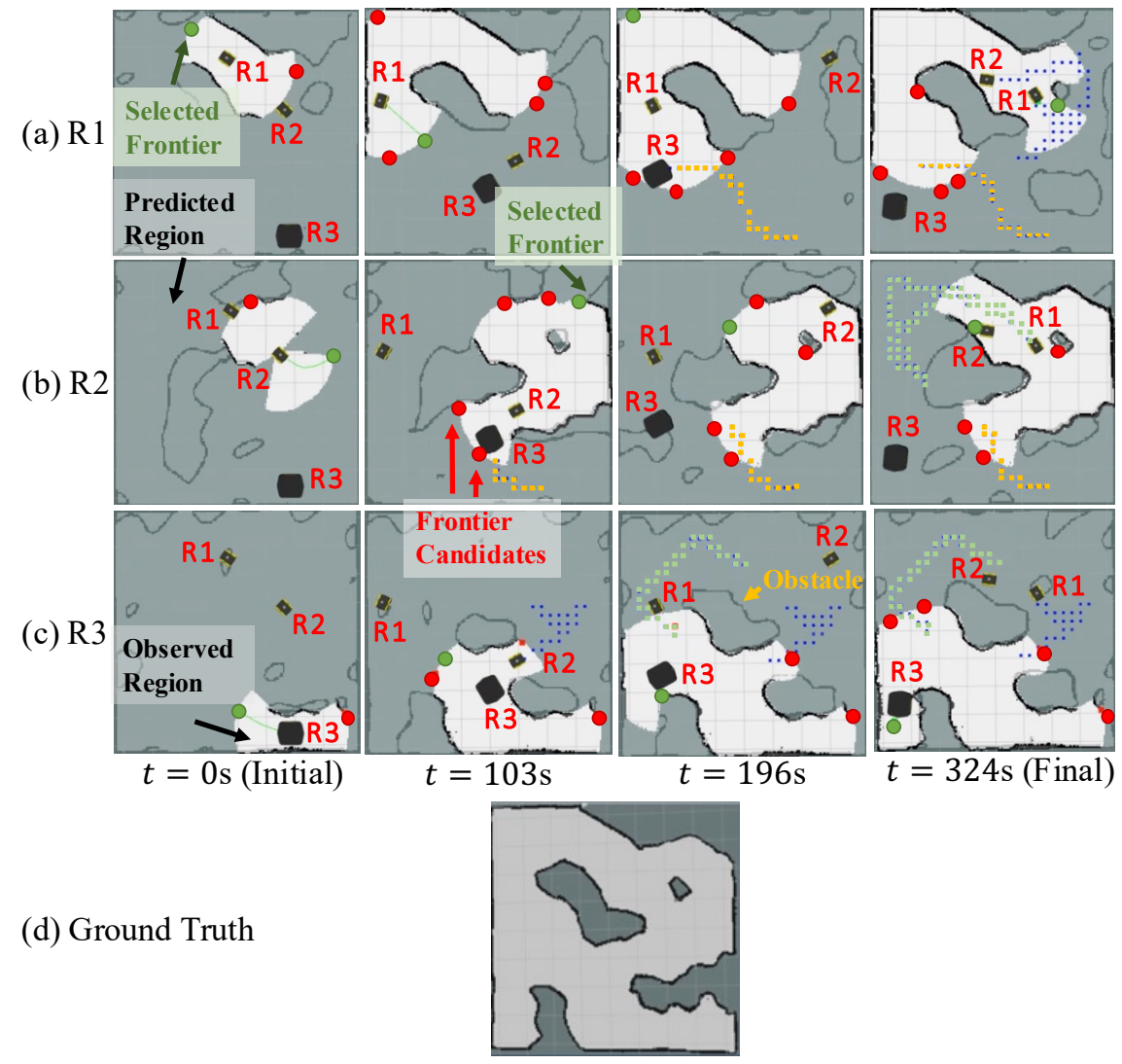

**Fig. 12.** (a) - (c) presents the map predictions at four distinct time steps for Robots 1-3: 0s (initial), 103s, 196s and 324s (final). Observed regions are in white and target regions for map prediction are in gray. The communicated trajectories between R1, R2 and R3 are represented by green, blue, and yellow dotted lines, respectively. Red circles represent frontier candidates, while green circles represent the selected frontier; and (d) represents the ground truth map.

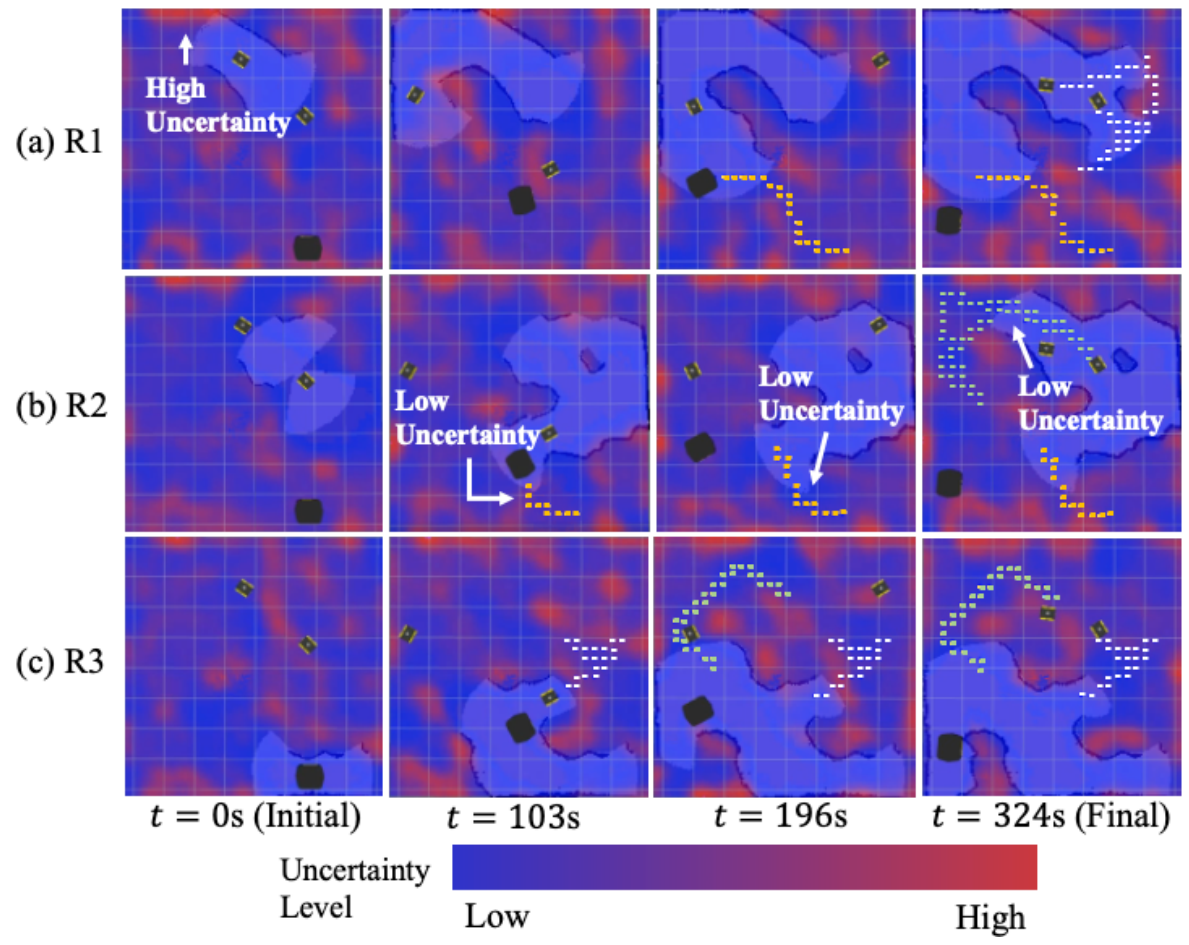

**Fig. 13.** (a) – (c) presents the confidence maps of Robot 1, 2 and 3 at time steps 0s, 103s, 196s and 324s. Red and blue represent high and low prediction uncertainty, respectively. White, green, and yellow dotted lines represent the communicated R1, R2, and R3 trajectories.

## IX. Conclusion

In this paper, we present a novel robot exploration with map prediction architecture called *4CNet-E*. Our main contributions include: 1) the first utilization of a consistency model in the development of a map prediction network for prediction of the spatial layouts in partially explored environments; 2) the unique application of contrastive learning for pre-training a trajectory encoder in order to consider both static and dynamic environment features; and 3) the introduction of a confidence network for map prediction for guiding robots towards areas of high uncertainty to improve map prediction accuracy within constrained energy budgets. Through extensive simulated comparison experiments, *4CNet-E* was shown to have better performance in terms of map prediction accuracy, and area coverage when compared to heuristic and learning-based methods. An ablation study validated individual design choices within *4CNet*, while a runtime analysis presented the computation cost of *4CNet* with respect to SOTA map prediction methods. Hardware experiments in both real natural outdoor environments, and unstructured indoor environments demonstrated the applicability and generalizability of *4CNet-E* to provide high quality map predictions, when compared to existing SOTA map prediction methods. Future work will include optimizing map prediction speed by reducing the number of time steps required to generate accurate map predictions using consistency models. This will allow for dynamic frontier generation, thereby, further improving the exploration efficiency.

## Acknowledgment

The authors would like to thank the following members of the ASBLab at UofT, Yuntao Cai, Haitong Wang, Daniel Choi, and Angus Fung for their invaluable discussions and assistance with the physical experimental set-up.